\documentclass[twoside,11pt]{article}

\usepackage{dnd}
\usepackage{times}
\usepackage{lipsum}
\usepackage{linguex}
\usepackage{subcaption}
\usepackage[usenames,dvipsnames]{xcolor}
\usepackage{bbding}
\usepackage{ulem}
\usepackage{booktabs}
\usepackage{colortbl}
\usepackage{tikz}
\usepackage{pgfplots}
\pgfplotsset{compat=newest}
\usepackage{circledsteps}

\newcommand{\sina}[1]{}
\newcommand{\sinameta}[1]{}

\firstpageno{1}

\begin{document}

\title{Implicit Causality-biases in humans and LLMs as a tool for benchmarking LLM discourse capabilities}

\author{\name Florian Kankowski \email florian.kankowski@uni-bielefeld.de \\
       \addr Bielefeld University/CRC 1646
       \AND
       \name Torgrim Solstad \email torgrim.solstad@uni-bielefeld.de \\
       \addr Bielefeld University/CRC 1646
       \AND 
       \name Sina Zarrieß  \email sina.zarries@uni-bielefeld.de\\
       \addr Bielefeld University/CRC 1646
       \AND
       Oliver Bott \email oliver.bott@uni-bielefeld.de\\
       \addr Bielefeld University/CRC 1646}
       
\editor{NA}
\submitted{NA}{NA}{NA}

\maketitle

\begin{abstract}%
In this paper, we compare data generated with mono- and multilingual LLMs spanning a range of model sizes with data provided by human participants in an experimental setting investigating well-established discourse biases. Beyond the comparison as such, we aim to develop a benchmark to assess the capabilities of LLMs with discourse biases as a robust proxy for more general discourse understanding capabilities. More specifically, we investigated Implicit Causality verbs, for which psycholinguistic research has found participants to display biases with regard to three phenomena:\ the establishment of (i) coreference relations (Experiment 1), (ii) coherence relations (Experiment 2), and (iii) the use of particular referring expressions (Experiments 3 and 4). With regard to coreference biases we found only the largest monolingual LLM (German Bloom 6.4B) to display more human-like biases. For coherence relation, no LLM displayed the explanation bias usually found for humans. For referring expressions, all LLMs displayed a preference for referring to subject arguments with simpler forms than to objects. However, no bias effect on referring expression was found, as opposed to recent studies investigating human biases.\label{firstpage}
\end{abstract}


\section{Introduction}

Despite recent successes of large language models (henceforth, LLMs), matching or exceeding human performance in wide array of tasks, it is still not clear to what extent they can be considered as models of human language \citep{milli2024} or human cognition \citep{mahowald2024dissociating}. Ever since the emergence of large foundation language models, there has been a great interest in gaining a deeper understanding of their general linguistic and cognitive capabilities, ranging from studies into syntactic or semantic processing in LLMs \citep{gulordava2018colorless, blevins2018deep, goldberg2019assessing, arefyev-etal-2020-always} to analyses of their arithmetic or logical reasoning abilities \citep{huang2022towards, webb2023emergent, wu2023reasoning, yuan2023well}. Since LLMs are trained on massive amounts of training data and learn huge amounts of parameters in their internal layers, it is often difficult to establish whether an LLM has ``really'' generalized a certain rule or piece of knowledge, or whether it memorized superficial token-level patterns \citep{weissweiler2023}. This challenge exists in particular in domains where linguistic knowledge is closely intertwined with non-linguistic aspects of cognition \citep{mahowald2024dissociating} such as, e.g., pragmatics and commonsense reasoning \citep{chang2024language}. %

To assess LLMs linguistic capabilities research has drawn from the methods of psychology, applying established experimental paradigms to language models. For example, \citet{ettinger2020bert} created a diagnostic suite for language models inspired by psycholinguistic tests, ranging over different linguistic disciplines. A study by \citet{hawkins2020investigating} compared human and model acceptability judgements for double object sentences, capturing the differences in their biases. Outside the realm of psycholinguistics, \citet{binz2023using} used vignette studies from cognitive psychology to gauge the cognitive abilities of GPT-3 \citep{brown2020language}. 
As the abilities and potential applications of language models widen, there has been a surge in benchmarking systems that either allow for a more sophisticated model evaluation \citep{kiela2021dynabench} or extrapolate far beyond NLP into tasks based on general reasoning, world-knowledge or domain-specific problem-solving \citep{srivastava2022beyond}. This reflects a general shift towards using LLMs as general-purpose problem solvers. However, fine-grained linguistic analysis of even highly advanced models can still be useful. For instance, although GPT-3 is capable of various NLP and problem-solving tasks \citep{zong2022survey,yang2022empirical,zong2023solving}, it still exhibits undesirable linguistic behaviour \citep{schuster2022sentence}. It should be noted that a lot of the existing research, especially in benchmarking, focuses on isolating specific phenomena without assessing the interconnectivity between linguistic tests on the one hand and their implications for broader reasoning capabilities on the other.

In this paper, we investigate whether LLMs exhibit human-like behaviour for a linguistic phenomenon called `Implicit Causality' (IC), which is widely researched in pragmatics and has connections to analyses of LLMs' general causal reasoning ability \citep{hong-etal-2024-large}. As such, IC is a cover notion for a number of properties associated with interpersonal verbs like \textit{fascinate} and \textit{admire} \ref{ex:intro}. Over the last fifty years, linguistic and psycholinguistic research has investigated how participants continue such sentences (with typical continuations in parentheses):

\ex.\label{ex:intro} \a. Mary fascinated Peter.\label{ex:intro:se} \HandPencilLeft \hfill{(e.g., \textit{She always came up with great suggestions})}
\b. Mary admired Peter.\label{ex:intro:es} \HandPencilLeft \hfill{(e.g., \textit{He was a great dancer})}

It has been shown that IC verbs display three different biases. First, a very strong coherence bias towards explanations over all other discourse relations taken together \citep{Kehleretal2008,SolstadBott2022}. Second, a coreference bias towards one of the arguments over the other in these explanations (\textit{Mary} in \ref{ex:intro:se} and \textit{Peter} in \ref{ex:intro:es} \citep[e.g.,][]{GarveyCaramazza1974,BrownFish1983,RudolphFoersterling1997}. Furthermore, depending on the verb type the coreference bias can be reversed if the discourse relation is changed from an explanation to a consequence relation \citep[][inter alia]{CrineanGarnham2006,Garnhametal2020ICons,Hartshorneetal2015,Stewartetal1998}.  Finally, it has also been observed that these preferred arguments have a tendency to be referred to by simpler morpho-syntactic forms than the disprerred ones \citep[e.g., pronouns instead of repeating the proper name, cf.][]{RosaArnold2017,WeatherfordArnold2021,BottSolstad2023,Dembergetal2023}. This set of interrelated biases together with the rich set of human data from carefully controlled continuation task experiments makes Implicit Causality a particularly well-suited phenomenon to set up a benchmarking system for investigating the causal discourse pragmatic abilities of LLMs.

Existing studies of IC in LLMs have focused only on the influence of IC on the establishment of coreferential relations, mostly in sentence continuations after \textit{because}. They have come to rather mixed conclusions as to whether LLMs exhibit such a bias \citep{kementchedjhieva2021john, davis2020discourse, davis2021uncovering}. A number of early studies even observed a complete lack of IC bias or even an inverse bias in LLMs for different languages \citep{upadhye2020predicting, zarriess2022isn}. However, the picture is different for recent larger LLMs, where strong human-like IC biases appear to emerge \citep{cai2023does}. Promising as these first findings on the coreference IC bias are, it is yet not clear whether they really speak in favor of human-like discourse pragmatic abilities. In human processing, the coreference bias depends on the coherence bias \citep{Kehleretal2008,SolstadBott2022}, and the form bias in turn depends on the coreference bias in tandem with other factors \citep{BottSolstad2023,Dembergetal2023,WeatherfordArnold2021}. In order to properly benchmark the general discourse capabilities of LLMs using Implicit Causality, all three biases as well as their interplay would ideally be analyzed together.

The central questions of this paper are (i) how human-like a number of LLMs perform on our benchmarking study comprising the whole set of Implicit Causality biases, and (ii) whether IC-related causal knowledge is contingent on the type of language model (mono- vs. multilingual) as well as on language model size, i.e. improves with increasing size. We analyze language model outputs for prompts that contain IC verbs, i.e. verbs known to exhibit a well-known coreference bias in the context of causal connectives (see Section \ref{sec:backgroundIC}). We thus go beyond previous studies that focused almost exclusively on specific types of causal prompts for these verbs and conduct behavioral assessments for all three types of biases. In addition, we extend our methodology to the study of contingency relations more generally by not restricting ourselves to \textit{because} prompts, but also study RESULT relations by prompting on \textit{and so}. By studying whether and how these different biases are orchestrated in LLMs, we hope to gain a deeper understanding of their ability to model Implicit Causality in particular and causal (discourse) connections in general.

To foreshadow our results, none of the models investigated in this study exhibit human-like causal discourse with respect to the coherence and the form biases. Nevertheless, apparent coreference biases were present in some of the larger models. So, we are mainly presenting a negative result. On the positive side, however, the study shows that we have to look beyond just the coreference bias to behaviorally assess LLMs understanding of Implicit Causality. Additionally, we have established a thorough benchmark for probing the discourse processing capabilities of LLMs, which can be used for assessing larger or more capable LLMs in the future.

This paper is structured as follows: In section \ref{sec:background}, we introduce the phenomenon at hand. In section \ref{sec:modelsetup}, we describe our process for model selection and data generation. Section \ref{sec:structure} lays out the structure of the procedure of our experiments that are discussed in sections \ref{sec:exp1} through \ref{sec:exp4}. In section \ref{sec:discussion}, the central findings are discussed and section \ref{sec:conclusion} concludes the paper. 

\section{Implicit Causality and Consequentiality as a test case for LLMs}
\label{sec:background}

As mentioned in the Introduction, particular aspects of Implicit Causality have already been investigated in previous research on the capabilities of LLMs. However, our investigation differs from those studies in (at least) two very important aspects. First, as we will discuss in detail in the next section, we compare different LLMs in terms of size and model type. Second, and equally important, we intend to establish a benchmarking system based on the whole range of phenomena associated with Implicit Causality. We argue that by focusing on coreferential biases, previous research on LLMs has investigated only one facet of a broader range of phenomena of relevance to the human construal of contigency relations. In the following, we will first introduce the phenomena associated with Implicit Causality to allow for a better contextualization of our study within the investigation of the capabilities of LLMs.

\subsection{Background:\ Implicit Causality and Consequentiality}
\label{sec:backgroundIC}

IC verbs have been extensively studied in psycholinguistics and a subject of recent interest in computational linguistics. Although the phenomenon of Implicit Causality was initially only investigated with respect to patterns of coreference in explanations, subsequent research has shown, as discussed briefly in the Introduction, that IC verbs are actually associated with three different, but interrelated biases. 

In general, IC verbs denote interpersonal relations, as with the stimulus-experiencer (experiencer-object verbs, e.g., \textit{annoy}, \textit{fascinate}) and  experiencer-stimulus (experiencer-subject verbs, e.g., \textit{admire}, \textit{hate}) subclasses of psychological verbs. IC verbs have been found to display three pragmatic biases in discourse, all of which will be of importance in this paper. These relate to preferences for establishing (i)  coreference relations, (ii) coherence/discourse relations, as well as (iii) for using specific anaphoric expressions in establishing coreference relations.

First, and most prominently, IC verbs induce a \textbf{next-mention coreference bias} to either the subject or the object in explanations, as has been shown in a number of production and comprehension tasks for several languages \citep{Au1986,BottSolstad2014,BottSolstad2021,BrownFish1983,Ferstletal2011,Goikoetxeaetal2008,Garnhametal2020ICons,GarveyCaramazza1974,Hartshorneetal2015,Hartshorneetal2013,RudolphFoersterling1997,SolstadBott2022}:

\ex.\label{explanation} \a. Mary fascinated Peter because \dots{} \HandPencilLeft{} she was very clever.
\b. Mary admired Peter because \dots{} \HandPencilLeft{} he was such a good dancer.

\noindent
In studies investigating these so-called Implicit Causality (henceforth, I-Caus) biases after ``\dots'' in \ref{explanation}, participants have been found to display strong next-mention biases:\ After the psychological stimulus-experiencer verb \textit{fascinate} participants provide continuations about the subject \textit{Mary} and after experiencer-stimulus verb like \textit{admire} they make primary reference to the object \textit{Peter}. Explanations aligned with the coreference bias are characterized as \textit{bias-congruent}, whereas continuations in violation of it, as in \ref{bias-incongruent} are said to be \textit{bias-incongruent}.

\ex. Mary fascinated Peter because he likes clever persons.\label{bias-incongruent}

The very same verbs have also been investigated for their next-mention bias following result/consequence connectives such as \textit{and so} \citep{Au1986,Commandeur2010,CrineanGarnham2006,Garnhametal2020ICons,Hartshorneetal2015,Stewartetal1998}:

\ex.\label{consequence} \a. Mary fascinated Peter and so \dots{} \HandPencilLeft{} he asked her for an interview.
\b. Mary admired Peter and so \dots{} \HandPencilLeft{} she asked him to give a talk. 

As with explanations, verb class-internal next-mention biases for these Implicit Consequentality biases (henceforth, I-Cons) are highly consistent. Interestingly, stimulus-experiencer (\textit{fascinate}) and experiencer-stimulus (\textit{admire}) verbs display reversed next-mention biases for I-Caus and I-Cons, a point to which we will return below.

A second property of these verbs is their \textbf{coherence bias}. When prompted to provide continuations after a full stop \ref{full_stop}, participants primarily provide explanations over other discourse relations \citep[around 60\% vs.\ 20-25\% for other (non-IC) verb classes, cf.][]{BottSolstad2014,BottSolstad2021,Kehleretal2008,SolstadBott2022,BottSchrumpfMichaelisSolstad2023,SolstadBott2023}. This extraordinarily high explanation bias with more explanations over all the other discourse relations taken together was consistently found for all types of IC verbs, psych verbs as well as agent-patient verbs, with no differences between verb types.

\ex.\label{full_stop} Peter admired Mary. \dots{}  \HandPencilLeft{} 
\a. Preferred:\ She was very clever \hfill (Explanation).
\b. Dispreferred:\ He decided to ask her for an interview. \hfill (Result)

A third bias relates to how fulfilling or violating next-mention biases affects the particular \textbf{form of the anaphoric expressions} used . Since proper names are generally assumed to be more complex than personal pronouns \citep[cf.\ e.g.][]{Ariel1990,Gundeletal1993}, one could assume that bias-incongruent continuations should also involve more complex forms \citep{Arnoldetal2013}:

\ex.\label{congruent_incongruent} \a. Mary fascinated Peter because \dots{} \HandPencilLeft{} she/\sout{Mary} was very clever. \hfill (bias-congruent)
\b. Mary fascinated Peter because \dots{} \HandPencilLeft{} {\sout{he}}/Peter likes clever persons. \hfill (bias-incongruent)

Although initial studies argued for a dissociation between referential predictability and referential form \citep{FukumuravanGompel2010,Kehleretal2008,KehlerRohde2013}, recent research has shown that next-mention biases do indeed bear an influence on referential forms, albeit conditioned on a number of factors \citep{WeatherfordArnold2021,BottSolstad2023,Dembergetal2023}. Thus, for instance, \citet{BottSolstad2023} found forced coreference to the object of IC verbs to be established by means of a significantly larger proportion of personal for bias-congruent objects than for bias-incongruent objects, see \ref{congruent_incongruent}.

As a final observation in support of the assumption that Implicit Causality consitutes a worthwhile case study for LLM capabilities, previous research has found coreference and coherence biases to exert influence on the prediction of linguistic input during the online processing of discourse \citep[cf.\ e.g.][]{KoornneefvanBerkum2006,vanBerkumetal2007,PyykkoenenJaervikivi2010,Garnhametal2020}. Thus, IC also appears to impact the online processing of language beyond the offline production biases discussed above. %

\begin{figure}[ht!]

\begin{center}
\scalebox{.8}{\begin{tikzpicture}
\draw node[thick] (SEMREP) {\textsc{semantic representation}};
\draw (node cs:name=SEMREP) +(0,-2.1) node (SYNREP) {\textsc{syntactic representation}};

\draw (node cs:name=SEMREP) +(0,-.9) node[fill=black!15!white] (SENT) {\textit{ Mary fascinated Peter}};

\draw (node cs:name=SEMREP) +(6.1,1) node [very thick,minimum width=3cm,text width=2.8cm] (EXPLTYPE) {\textbf{Relation}};
\draw (node cs:name=EXPLTYPE) +(0,-2) node [very thick,minimum width=3cm,text width=2.8cm] (REFTYPE) {\textbf{Referent}};
\draw (node cs:name=EXPLTYPE) +(0,-4) node [minimum width=3cm,text width=2.8cm] (ANAFORM) {Anaphoric form};

\draw (node cs:name=EXPLTYPE) +(1.8,0) node [circle,draw] {{\footnotesize{1}}};
\draw (node cs:name=REFTYPE) +(1.8,0) node [circle,draw] {{\footnotesize{2}}};
\draw (node cs:name=ANAFORM) +(1.8,0) node [circle,draw] {{\footnotesize{3}}};

\draw [->,line width=1mm] (SEMREP.east) to[out=50,in=180] node[midway,rotate=15,above] {predicts} (EXPLTYPE.west);
\draw [->] (SYNREP.east) +(0,0) to[out=30,in=180] node[midway,rotate=30,above] {predicts} (REFTYPE.west); 
\draw [->] (SYNREP.east) +(0,0) to[out=-30,in=180] node[midway,rotate=-30,below] {predicts} (ANAFORM.west); 
\draw [->,line width=.5mm] (node cs:name=EXPLTYPE) +(-.7,-.5) to node [rotate=270,above] {predicts} +(-.7,-1.5);
\draw [->] (node cs:name=EXPLTYPE) +(-.7,-2.5) to node [rotate=270,above] {predicts} +(-.7,-3.5);

\end{tikzpicture}}

\end{center}

\caption{Three-level model of discourse expectations associated with IC%
}\label{figure:model}
\end{figure}
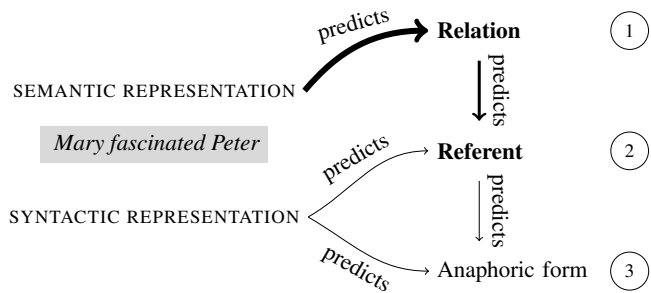

As mentioned above, a key feature of Implicit Causality is the fact that the biases induced by it are not coincidental, but assumed to be inextricably related, at least in human processing, which makes the phenomenon so attractive to study in LLMs. To this end, it is important to understand exactly how the different IC effects interact. We propose to model the interconnection between the three biases discussed above as illustrated in Figure \ref{figure:model}.

On the top-most level\  \Circled{{\footnotesize{1}}}, an expectation for a particular type of explanation is evoked \citep{BottSolstad2014,BottSolstad2021,SolstadBott2022,SolstadBott2023}. Once a discourse relation is determined at level\  \Circled{\footnotesize{1}} by means of a connective such as \textit{because} or \textit{and so}, this influences the expectation for a particular referent to occur at level\  \Circled{\footnotesize{2}} \citep[see][and much subsequent research]{Ehrlich1980}. Finally, the particular choice of a referent at level\  \Circled{\footnotesize{2}} influences expectations for a particular referential form at level\  \Circled{\footnotesize{3}}. These effects are predicted to be the most subtle of the three biases, as they constitute top-down, low-level predictions from discourse to morpho-syntax (i.e., levels\  \Circled{\footnotesize{1}} through\  \Circled{\footnotesize{3}}), predicted to display more elusive effects by \citet{PickeringGambi2018}.

\subsection{Implicit Causality as a benchmark system for LLM systems}

In the above, we argued that Implicit Causality constitutes an expectation-based phenomenon impacting language processing at the level of coreference, coherence, and form of expression. It thus seems to be particularly suited for comparing human biases with computational models intended to mimic the linguistic behaviour of humans. What makes these biases a particularly interesting test case for the discourse competences of LLMs, are their interrelations as discussed with respect to Figure \ref{figure:model} and intricate interactions with other linguistic factors such as argumenthood and linearity for referential forms. 

Firstly, as just outlined, the coreference bias is strongly interrelated with the coherence bias, since the particular coreference majorly interacts with the coherence relation \citep{Kehleretal2008,SolstadBott2022}. Secondly, the three biases differ in their relative strength and it is thus interesting to see whether they can be reproduced in LLMs. Thirdly, next-mention biases strongly interact with other factors in particular with respect to the anaphoric form bias. Interactions exist with the grammatical function of the antecedent and pragmatic constraints on particular anaphoric forms \citep[see][for discussion]{BottSolstad2023}.

Thus, in order to be able to claim that LLMs have grasped a human discourse bias such as the one associated with Implicit Causality, it does not suffice to investigate coreference biases . This opens up the interesting possibility that LLMs may be able to predict next word predictions necessary for next-mention coreference biases, but fail to grasp other aspects of importance to make discourse relation predictions on level\   \Circled{\footnotesize{1}} in Figure \ref{figure:model}. An even stronger statement can be made in regard to generalized linguistic ability: In order to display human-like behaviour with respect to the three IC biases presented here, an LLM would need to possess competency in a number of related domains, rendering IC a suitable proxy for a language model's broad discourse processing capabilities.

\subsection{IC effects in language models}

Thanks to its strong foundation in psycholinguistic research, the phenomenon of IC has already been established in experimental computational linguistics for studying discourse effect sensitivities in language models. However, existing papers exclusively focused on the coreference bias of implicit causality verbs neglecting coherence and form biases. Furthermore, existing studies haven't compared LLM's perfomance related to different next-mention coreference biases such as implicit causality and implicit consequentiality. This means that the current research was unable to capture a bigger picture of the LLMs' abilities for processing IC and their discourse capabilities more generally.

The first study in this area was performed by \cite{upadhye2020predicting}, who used token probabilities in language models conditioned on implicit causality prompts to measure the models' sensitivity to IC biases for anaphoric coreference. They observed biases in GPT-2 L \citep{radford2019language} that were either significantly weaker or even inverse to human biases. It should be noted, that they only considered personal pronoun coreferences in their bias calculations. \cite{kementchedjhieva2021john} used a similar experimental setup to show that the IC biases of language models do not conform well to the human standards, with GPT-2 M exhibiting a general coreference bias towards the object of prompts that superceded any possible IC effect. On the contrary, \cite{davis2020discourse} did find a robust significant IC effect for GPT-2 XL in a similar coreference task, where bias-congruent pronouns were associated with significantly lower surprisal than bias-incongruent ones. However, they argued that bias-congruent model behaviour is mostly a product of surface-level behaviour and not indicative of a model-internal understanding of the associated discourse-effects. They motivated this argument by means of probing, showing that IC verbs only weakly modulate the internal representations of anaphoric pronouns.

An investigation by \cite{zarriess2022isn} into IC biases of German instatiations of GPT-2 and BERT \citep{devlin2019bert} painted a similar picture, with the models exhibiting weak IC biases, if any. In fact, they observed that changing the gender or names of the referents in the prompts had a greater impact on coreference preference than IC verb type, suggesting completely incoherent internal representations. Similar to \citet{kementchedjhieva2021john}, they found a general object bias in the model predictions, which is generally incompatible with human behaviour. \citet{davis2021uncovering} hypothesize that the mismatch between model and human bias may be a result of competing linguistic constraints. Under this assumption, the models may actually learn internal representations for IC verbs, but not exhibit the corresponding bias, because it is masked by another constrained. They motivated this hypothesis by showing that the mismatch between human and LLM preferences can be diminished through appropriate finetuning that targets competing constraints.

More recent research investigated not only the IC bias itself, but also the ability of models to generate contextually plausible bias congruent continuations (\citet{sieker2023beyond}, \citet{huynh2022implicit}). As part of a larger study on the linguistic capabilites of ChatGPT, \citet{cai2023does} showed that the model is not only sensitive to IC coreference effects, but exhibits a much stronger bias than humans.

\section{}

\section{Language Model  Setup}
\label{sec:modelsetup}

This section explains our rationale for selecting the models we used in our study as well as well as the procedure for generating the test data.

\subsection{Model selection}
Or focus in this paper lies with the developed methodology for investigating IC effects in LLMs, not with the performance of any given model. As such, we performed our experiments on a range of different models in order to give a broad overview of the competence of contemporary LLMs. We made two significant carveouts in our selection process:

In the interest of open science, we only used publicly available white-box models, ignoring LLMs and generative AI services that are locked behind a paywall or that can only be prompted via API without access to the full weights. In particular, all the models used were fetched from the huggingface hub \citep{jain2022hugging}. We also restricted ourselves to small to medium-sized models that can be run on a single high-end consumer GPU with full floating point precision. In the future, we also want to consider larger models that are generally of more interest for use in application or as chatbots.

In this study, we only consider pretrained foundation models. We do not include LLMs that have been finetuned for some downstream task. In particular, we do not include any instruction-tuned or dialogue/chat systems. Instruction tuning has been observed to enhance model behaviour when it comes to typical NLP and general purpose tasks \citep{ouyang2022training}, but its effects on a model's general linguistic capabilites and language use behaviour are still unclear. In the context of IC, \cite{cai2023does} indicates that dialogue models align more with human behaviour, but \citet{gao2023roles} suggest that instruction-tuning has a negative, if slight effect on language perception. With the growing importance and presence of instruction and dialogue models, investigating their linguistic behaviour merits its own dedicated study, which we leave to future research.

Aside from these points, the following considerations in model selection were made in the interest of observing their significance in our experimental data:

\begin{itemize}
    \item \textbf{mono- vs. multilingual}: Although many modern state-of-the-art generative LLMs are inherently multilingual, this is not necessarily the case. Research on the performance difference of monolingual and multilingual models on NLP tasks is conflicting \citep{velankar2022mono,feijo2020mono}, so we decided to test both monolingual German models as well as multilingual models with a sufficiently large (top 5) representation of German in their training data. In regard to IC, the constraint-based mismatching mechanism hypothesized by \cite{davis2021uncovering} might apply very differently to multilingual models, which makes it worthwhile to study both groups.
    \item \textbf{model size}: Larger LLMs tend to perform better than smaller ones, both in terms of linguistic capability as well as downstream tasks (\citet{warstadt-etal-2020-learning}, \citet{gao2023roles}). However, some abilities -- like advanced logical reasoning -- emerge only on very large models while others -- like mastery of basic grammar -- are robustly present even on small models \citep{eldan2023tinystories, zhang-etal-2021-need}. We are interested in investigating whether the three IC-related effects tested in this paper emerge similarly in models or whether there exists a hierarchical representation, wherein the emergence of one effect requires the presence of another. To this end, we included model families in our experiments that come in different sizes.
\end{itemize}

With these considerations, the following models were chosen to be studied:

\begin{itemize}
    \item \textbf{GPT-2}: GPT-2 has been extensively studied in the context of Implicit Causality, showing a general insensitivity to IC effects under some conditions \citep{kementchedjhieva2021john,zarriess2022isn} and a significant, if small sensitivity to IC effects under others \citep{davis2020discourse,huynh2022implicit}. For this reason, it serves as a great baseline. We use a German instantiation of GPT-2 by \cite{stefanit2021gpt2large}.
    \item \textbf{mGPT}: mGPT \citep{aiforever2022mgpt} is a popular multilingual model utilizing a GPT-3-like architecture with a top-2 German representation in its training data.
    \item \textbf{XGLM}: The XGLM family of models \citep{facebook2021xglm} is a set of autoregressive multilingual LLMs trained with the goal of generalizing cross-linguistic model ability. We use the whole family of 564M, 1.7B, 2.9B, 4.5B and 7.5B models.
    \item \textbf{German BLOOM}: In order to be able to capture effects of model size in a monolingual setting, we include a family of monolingual German BLOOM models, consisting of a 350M model trained from scratch \citep{malteos2022bloom350} as well as a 1.5B and a 6.4B version \citep{malteos2023bloomclp}, both trained using CLP-transfer-learning \citep{ostendorff2023efficient} on the original BLOOM model \citep{bloom}.
\end{itemize}

\subsection{Prompting the models}
Previous work on IC effects in LLMs mostly used model output distribution, conditioned on an IC prompt, as a measure of coreference bias by comparing the probability of a congruent personal pronoun with the probability of an incongruent one (\citet{upadhye2020predicting}, \citet{kementchedjhieva2021john}). Instead, we opt for generating full sentence continuations and annotating the model outputs based on the relevant criteria, as one would do in a human language production experiment. The reasoning for this is threefold:

In German, the subject of an embedded clause need not be realised immediately after the connective, necessitating a lookahead of multiple tokens. Similarly, some of the chosen models need multiple tokens to generate some of the possible anaphoric forms (like first names) for coreferences.

In the same vein, a full continuation is necessary to investigate coherence effects of a generation or to judge sentence grammaticality (which can still be a problem with some models). 

Thirdly, generating full continuations allows us to explore the model outputs directly for futher analysis -- like investigating the form-effects or multiple anaphora in the model output. A fully annotated dataset also allows us to perform similar statistical analysis as is done with human data, aiding in comparability between the two.

Generating full continuations requires a decoding scheme. Selecting an adequate decoding method is a non-trivial task, as different decoding methods have been shown to produce vastly different model behaviour and output quality \citep{holtzman2019curious}. Notably, the quality and desirability of a decoding method is heavily task-dependent \citep{shi2024thorough} and can thus not be determined a priori. As a comparison of different decoding methods would exceed the scope of this paper, we restrict ourselves to one decoding style. \cite{sieker2023beyond} studied the acceptability and naturalness of machine-generated continuations to IC prompts as rated by human annotators using three different decoding methods. Following their results, we use diverse beam search \citep{vijayakumar2016diverse} with a beam size and beam group size = 10 and a diversity penalty $\lambda = .6$. 

We choose not to include any context in the prompt, like instructing the model about its completion task or giving information about the nature of the experiment, or enriching the short sentence fragment prompt with a surrounding story. Instead, we present the prompt as-is, letting the models freely generate continuations. The reasoning for this is that sentence continuation is the natural task for autoregressive foundation models and thus does not necessitate any context or instruction. As such, this makes the setup analogous to the sterile environment in human psychological trials where the presence of outside confounding stimuli is kept to a minimum.

\subsection{Data annotation}

Unlike for the human data presented here, the large amounts of data generated by the language models cannot feasibly be annotated by hand. Instead, we used a semi-automatic annotation scheme for labeling the generated data. Our automatic annotator parses the generated continuations with spaCy \citep{spacy2} and then traverses the tagged output and the dependency structure to extract the information that is relevant to each experiment. With this, a correctly labeled annotation can be created, as long as spaCy correctly identifies the grammatical features of the model's generation. Since the annotation differs between experiments, we describe the relevant annotation process in the \textsc{Data selection and annotation} subsection of sections \ref{sec:exp1} through \ref{sec:exp4}.

\section{Structure of the Study}
\label{sec:structure}

The test bed in our assessment of the LLMs' performance on the three biases -- coreference, coherence, and referring expressions -- follows a similar logic for all three biases. In the introduction to each experiment, we present what may be characterized as ``gold standard'' results from our own previous research involving human participants (coreference and coherence:\ \citeauthor{SolstadBott2022} \citeyear{SolstadBott2022}; referring expressions:\ \citeauthor{BottSolstad2023} \citeyear{BottSolstad2023}). That research involved psycholinguistic experiments that were conducted and analysed according to present date standards. Moreover, the experiments were all conducted within the same language German, using the same selection of verbs and comparable groups of participants. 

The data for the LLMs were sampled using the same experimental designs as in the gold standard experiments (with slight modifications where required due to the task at hand). In a next step, the human and LLM data were all annotated using the same semi-automatic annotation procedure to allow for a better comparison of LLM performance to the human data. Finally, descriptive and inferential statistics are presented that focus on the major findings from the human data, that is, investigating whether what was found for humans can also be found for individual LLMs.

In the next four sections, we will present the results from four experiments investigating next-mention coreference biases (Experiment 1), coherence biases (Experiment 2) and the potential influence of I-Caus and I-Cons next-mention biases on the choice of referring expressions (Experiments 3 and 4, respectively).

\section{Experiment 1:\ Coreference Bias}
\label{sec:exp1}

\citet{SolstadBott2022} investigated the I-Caus and I-Cons biases of stimulus-experiencer and experiencer-stimulus verbs in German, which were found to be reversed within verb classes for the two bias types. For instance, whereas the \textit{stimulus-experiencer} verb \textit{faszinieren} `fascinate' displays an I-Caus (\textit{because}) subject bias, it has an I-Cons (\textit{and so}) object bias. Accordingly, \citet{SolstadBott2022} elicited continuations manipulating the factors \textsc{verb class} (\textit{stimulus-experiencer} and \textit{experiencer-stimulus} verbs such as \textit{faszinieren} `fascinate' and \textit{bewundern} `admire' and \textsc{bias type} (\textit{I-Caus} vs.\ \textit{I-Cons}; as manipulated by the connectives \textit{weil} `because' and \textit{sodass} `and so', respectively). In addition, they included \textsc{gender order} of the proper name antecedents in the prompt (\textit{female subject-male object} vs.\ \textit{male subject-female object}) as a counter-balancing factor. Accordingly, 40 prompts with 20 stimulus-experiencer and 20 experiencer-stimulus verbs were constructed in four conditions:%
\footnote{In \citet{SolstadBott2022}, \textsc{verb class} was manipulated within-items based on the semantic relatedness of verbs. In the present study, however, this factor is treated as between-items. In \citet{SolstadBott2022} an additional analysis comparing these two design variants found no significant differences.}

\ex. \textbf{Stimulus-experiencer verb \textit{faszinieren} `fascinate'}
\a. \textit{Maria faszinierte Peter, weil \dots}  \HandPencilLeft \hfill (I-Caus; fem.-masc.)\\
`Mary fascinated Peter because \dots'
\b. \textit{Peter faszinierte Maria, weil} \dots \HandPencilLeft \hfill (I-Caus; masc.-fem.)\\
`Peter fascinated Mary because \dots'
\b. \textit{Maria faszinierte Peter, sodass \dots} \HandPencilLeft \hfill (I-Cons; fem.-masc.)\\
`Mary fascinated Peter and so \dots'
\b. \textit{Peter faszinierte Maria, sodass \dots} \HandPencilLeft \hfill (I-Cons; masc.-fem.)\\
`Peter fascinated Mary and so \dots'

\ex. \textbf{Experiencer-stimulus verb \textit{bewundern} `admire'}\label{ex:esitem}
\a. \textit{Emma bewunderte Karl, weil \dots}  \HandPencilLeft \hfill (I-Caus; fem.-masc.)\\
`Emma admired Karl because \dots'
\b. \textit{Karl bewunderte Emma, weil} \dots \HandPencilLeft \hfill (I-Caus; masc.-fem.)\\
`Karl admired Emma because \dots'
\b. \textit{Emma bewunderte Karl, sodass \dots} \HandPencilLeft \hfill (I-Cons; fem.-masc.)\\
`Emma admired Karl and so \dots'
\b. \textit{Karl bewunderte Maria, sodass \dots} \HandPencilLeft \hfill (I-Cons; masc.-fem.)\\
`Karl admired Emma and so \dots'

52 participants provided a total of 2,080 continuations that were manually annotated. Beyond general sensicality, it was coded whether the continuations included an anaphoric expression and whether this expression referred to the subject, the object, both arguments (e.g., \textit{they}) or neither. Only continuations containing an anaphora making unique reference to one of the referents (e.g., \textit{sie}/\textit{er} `she/he' or \textit{Emma}/\textit{Karl} for the prompts in \ref{ex:esitem}) were included in statistical analysis.

\begin{figure}[ht]
\centering
\includegraphics[width=.8\textwidth]{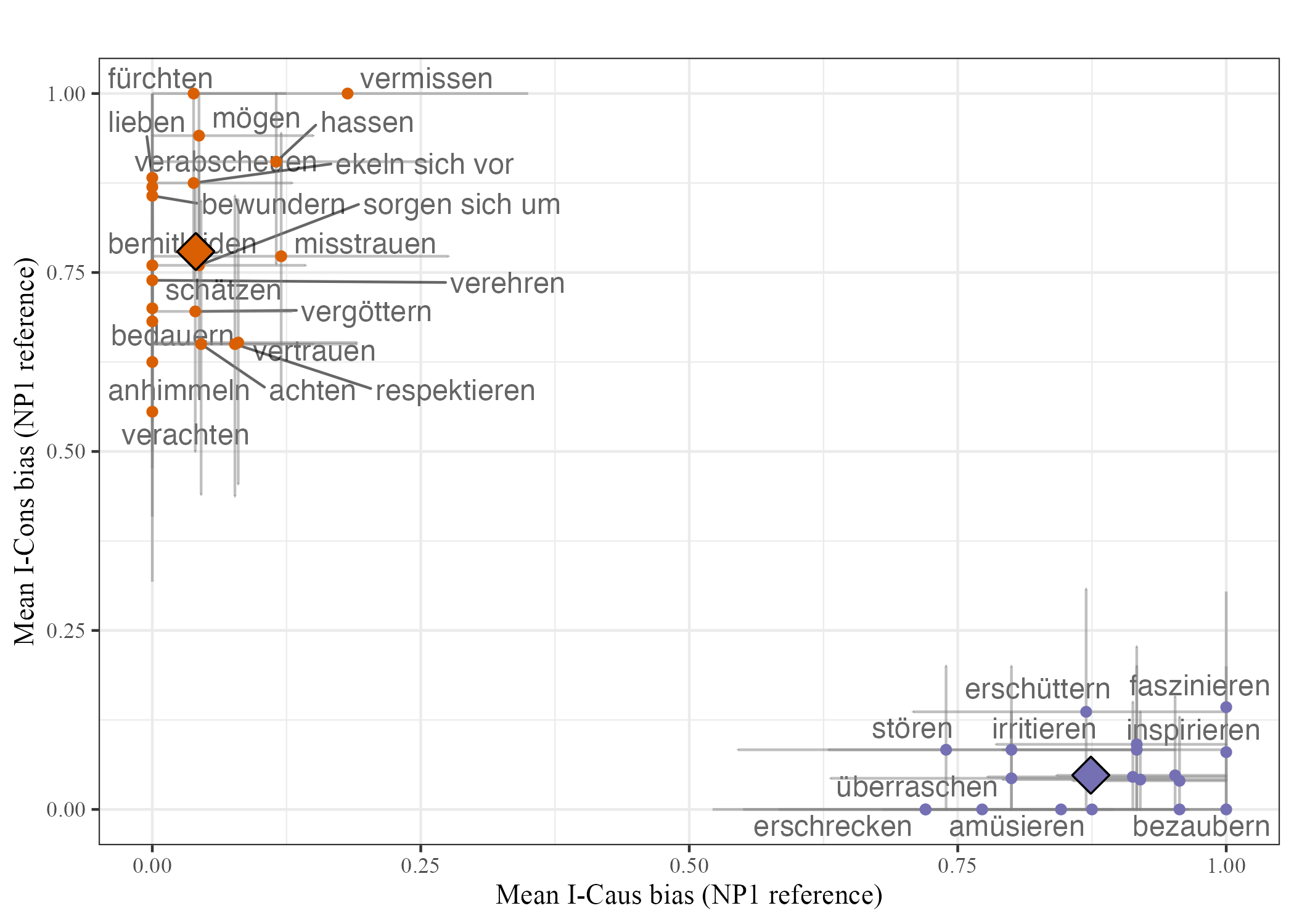}
\caption{Individual I-Caus and I-Cons biases for stimulus-experiencer and experiencer-stimulus verbs from \citeauthor{SolstadBott2022}'s \citeyearpar{SolstadBott2022} study. I-Caus bias is plotted on the $x$ axis and I-Cons bias on the $y$ axis (1 = coreference to the subject, 0 = coreference to the object) . The diamond-shaped points represent the average I-Caus and I-Cons biases for the two verb classes, respectively.}\label{fig:coref_gold_standard}
\end{figure}

In Figure \ref{fig:coref_gold_standard}, the individual I-Caus and I-Cons biases of the verbs tested in \citet{SolstadBott2022} are plotted. As can be readily seen, the verb classes form two distinct clusters. While stimulus-experiencer verbs (lower right corner) have an I-Caus bias towards the subject and an I-Cons bias towards the object, experiencer-stimulus pattern in the exact opposite direction for the two bias types. Statistical analysis revealed an almost perfect negative correlation between the two verb classes (Pearson's $r(37) = 0.94, p < .001$).%
\footnote{Concerning the df value reported here, it may be noted that one verb, the experiencer-stimulus verb \textit{gefallen} `please' was removed from the analysis \citep[for details, see][]{SolstadBott2022}\label{fn:df_gefallen}.} %
A logit mixed-effects regression model was computed predicting coreference to the subject referent as a function of the centered predictors \textsc{verb class} and \textsc{bias type} as well as random intercepts of participants and items. The analysis revealed the expected crossover interaction between \textsc{verb class} and \textsc{bias type} ($\chi^2(1) = 1336.2, p<.001$). In comparing the human and LLM data, we will primarily focus on this interaction.

\subsection{Methods}

The prompts for the LLM generation used the same design as for the human participants just reported. There were a few notable differences, however. First, only 19 (instead of 20) verbs from each verb class were used, since \citet{SolstadBott2022} excluded one item (consisting of a pair of verbs, see footnote \ref{fn:df_gefallen}) from statistical analysis as it turned out to display an unexpected behaviour.

Second, we used a different set of 40 female and 40 male first names optimally suited for LLMs to process the referents as unambiguously male or female:\ To this end, we used a list of the top 100 most common contemporary German first names, filtering out those that were deemed to be potentially ambiguous. Among the rest of the names, we removed those for which 5\% or more gender-incongruent anaphoric expressions (e.g., \textit{Maria} \dots{} \textit{er} `he' \dots) were observed for simple intransitive prompts using German GPT-2. 80 of the remaining names were paired such that all pairs were tested in all four conditions for the 38 verbs, resulting in a total of 6,080 continuations for each language model.

\subsection{Data annotation and selection}

The procedure for annotating the data ensured that every continuation submitted to statistical analysis could be assigned a syntactic parse . No other criteria for acceptability were applied. 

The automatic annotation process starts by parsing the generated continuations with spaCy to identify the subject in the continuation, if any. With this, the annotator can identify both the anaphoric coreference as well as its form. It is important to note that we only annotate coreferences if they are the subject of the continuation, as non-subject coreferents cannot be disambiguated by syntax alone.

To show that this annotation scheme still reliably labels coreference occurences, we calculated the inter-annotator-agreement between the manual and automatic annotation for the experiments performed in \citet{SolstadBott2022}, yielding $\kappa = .85$ and $\kappa = .996$ for coreference labeling in experiments 1 and 2, respectively. One caveat with this comparison is that the human annotators only created labels for continuations that were deemed sensible, while the automatic annotator considers all grammatical\footnote{Here, any sentence is considered grammatical if the spaCy parser is able to construct a full dependency structure for it. This also includes some sentences that are technically ungrammatical.} continuations. Restricting the agreement calculations to those sensible sentences increases the agreement score for experiment 1 to $\kappa = .89$.

Moreover, to allow for comparison with the data from \citet{SolstadBott2022}, the data entered into statistical analysis only included continuations in which the first anaphoric expression referred to either the subject or object argument in the prompt, as opposed to both or neither argument. Across LLMs, this led to the exclusion of between 5.9\% and 26.3\% of the data.
\subsection{Statistical analysis}\label{sec:analysis:coref}

Analysing the remaining continuations, we fitted a mixed-effects binomial logistic regression model for each set of data (human and LLMs alike) with the lme4 package \citep{Batesetal2015} in R (version 4.3.1). The binomial dependent variable coded whether coreference was established to the subject or object argument of the prompt. The model included the interaction between \textsc{verb class} and \textsc{bias type} in addition to the main effect of \textsc{gender order}, by-verb random slopes for \textsc{gender order} and \textsc{bias type} and their interaction in addition to a random intercept for verbs.%
\footnote{The greater number of productions allowed us to use more elaborate random effects structures than that employed for the human data in \citet{SolstadBott2022}. It should be noted that the random effects differed slightly in the analysis of human and LLM data. For humans, we also included a random intercept for participant, which could not be used for LLMs, where each continuation may be considered to be independent of all other continuation.} %

All factors in the model were centered. The maximal model just described was simplified in line with the findings in \citet{SolstadBott2022} by subsequently removing fixed effects. Model improvement was assessed by performing likelihood ratio tests of a model with the effect in question against a model without this effect. As mentioned above, the human data is characterized by a cross-over interaction between \textsc{verb class} and \textsc{bias type}. Consequently, the model involving the fixed effects \textsc{verb class} and \textsc{bias type} and their interactions was first compared with a model with these effects as main effects only. If the interaction turned out to be significant, the data were subsetted according to \textsc{bias type} (I-Caus and I-Cons, respectively) for which we compared models with \textsc{verb class} as the only fixed effect with ``intercept only'' models. The statistical procedure was implemented as loop, subjecting all (human and LLM) data to the same procedure. In Table \ref{inferential:coreference}, the corresponding $\chi^2$ values are reported along with their significance levels for human data and all LLMs.

For the I-Caus and I-Cons biases of individual verbs, we estimated 95\% confidence intervals for I-Caus and I-Cons biases by applying non-parametric bootstrapping \citep{Efron_Tibshirani_86} with the bootstrapping function from R’s \texttt{boot} package \citep{CantyRipley2024}. All analyses and data are publicly available in an OSF repository (\textcolor{red}{not yet published}).

\subsection{Results and Discussion}

\begin{figure}[ht!]
\begin{center}
\includegraphics{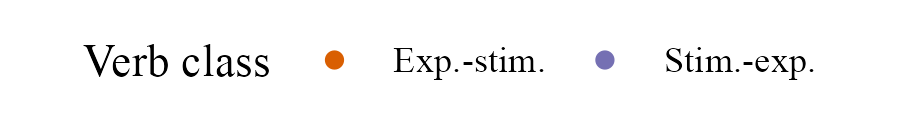}
\end{center}
\begin{subfigure}{0.32\textwidth}
\fcolorbox{white}{black}{
\includegraphics[width = .96\textwidth]{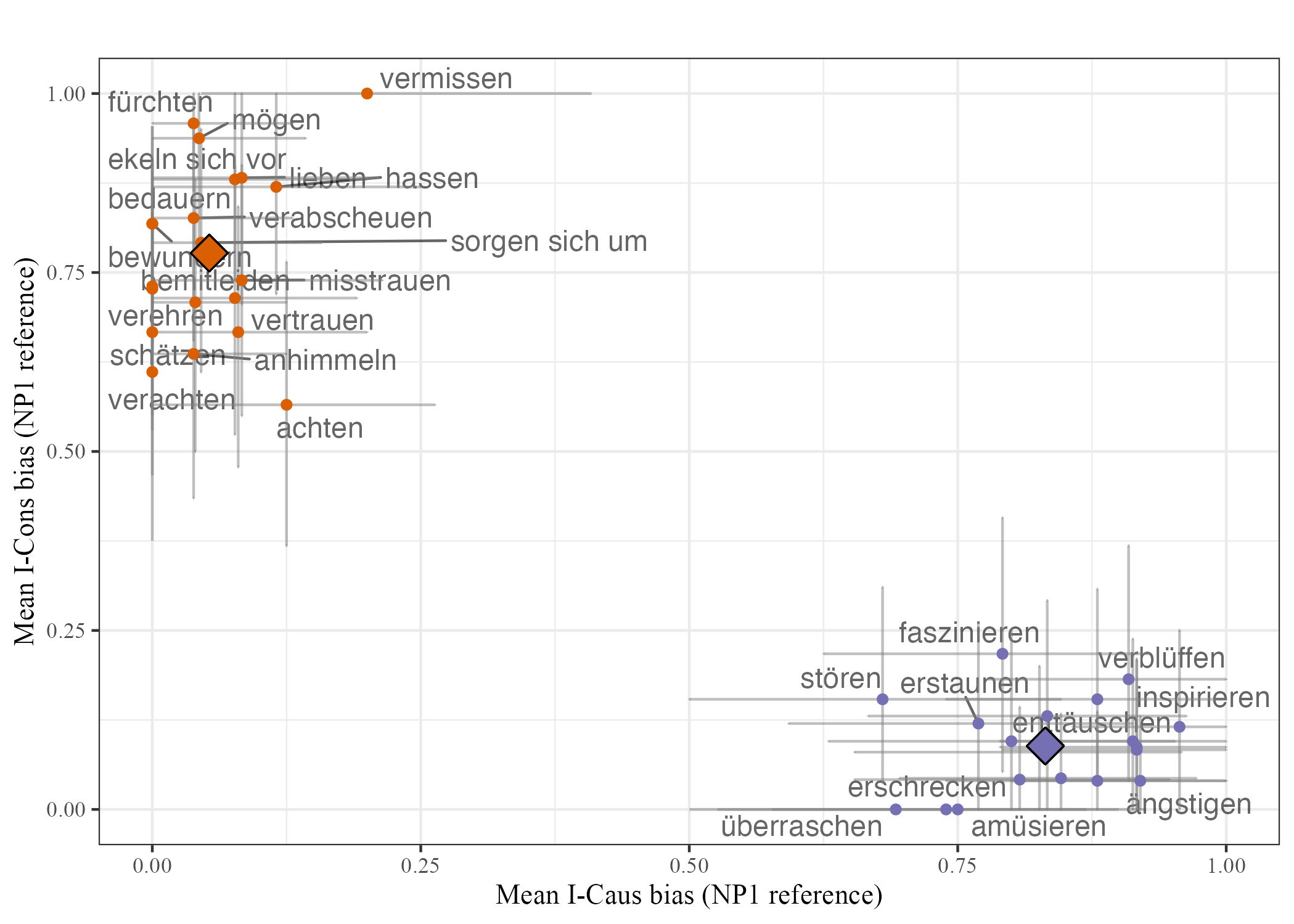}}
\caption{Human data (machine coded)}\label{coref_humans}
\end{subfigure}
\hspace{0pt}
\begin{subfigure}{0.32\textwidth}
\includegraphics[width = \textwidth]{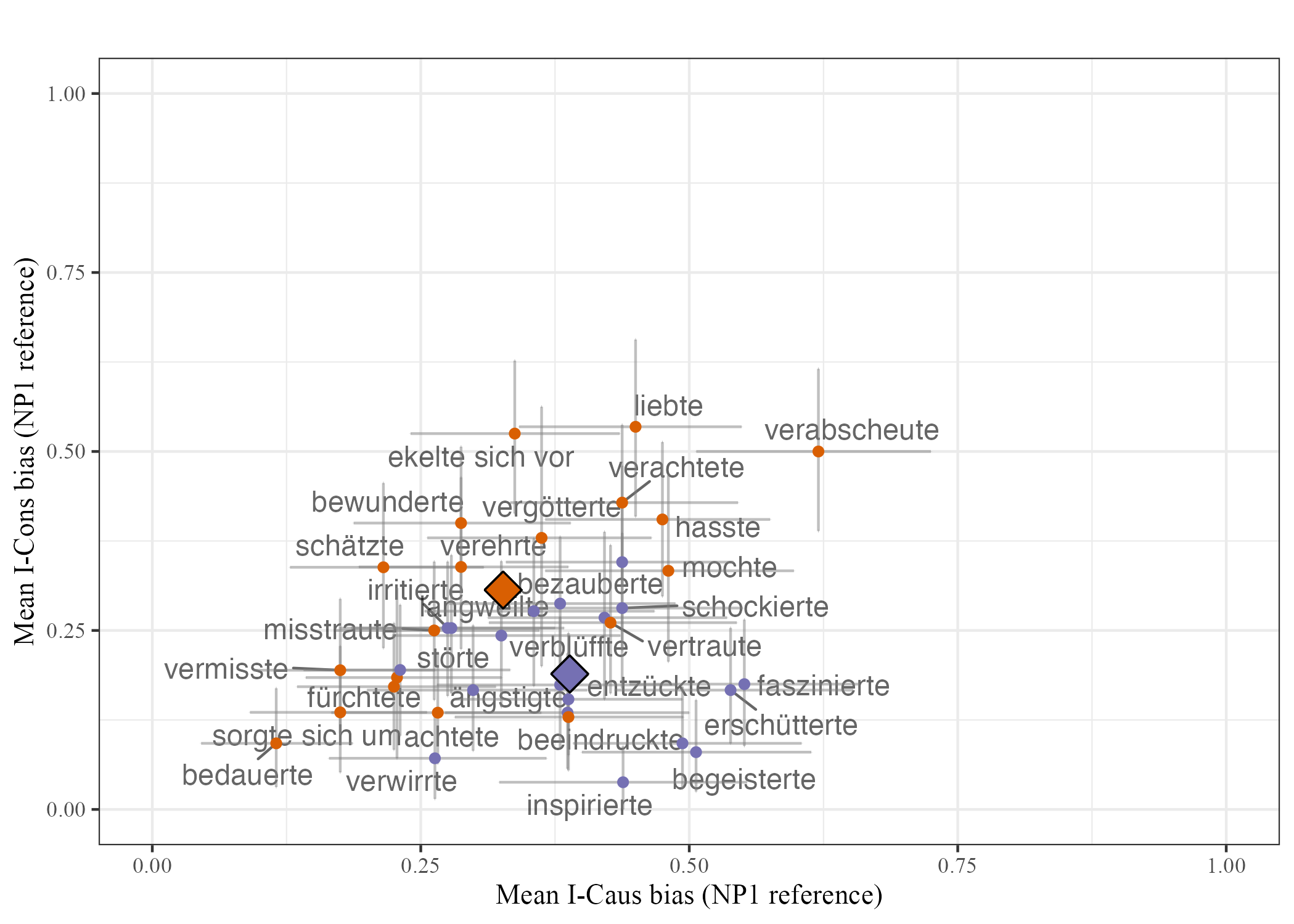}
\caption{German GPT-2}\label{coref_gpt2}
\end{subfigure}
\begin{subfigure}{0.32\textwidth}
\includegraphics[width = \textwidth]{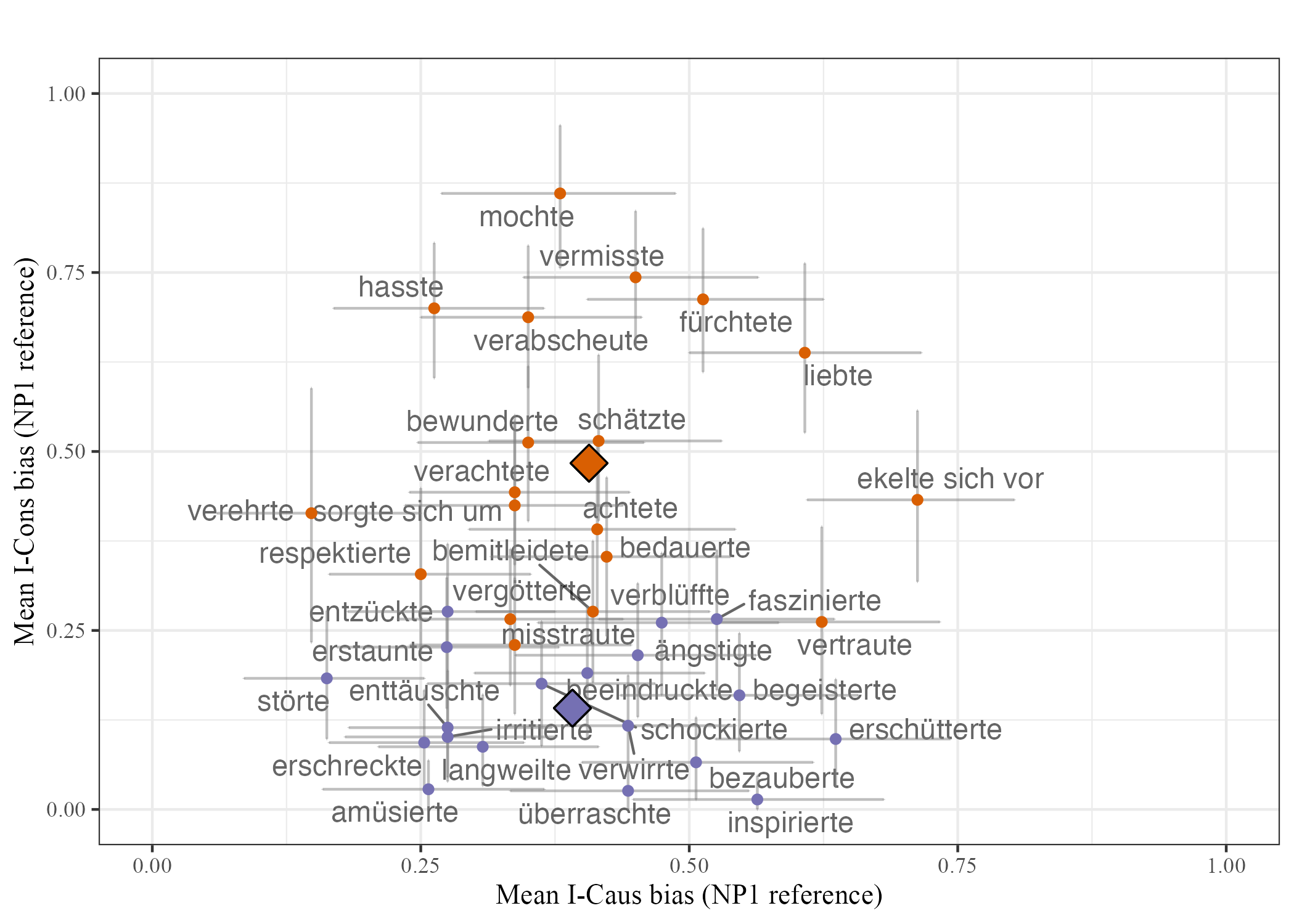}
\caption{mGPT}\label{coref_mgpt}
\end{subfigure}

\begin{subfigure}{0.32\textwidth}
\includegraphics[width = \textwidth]{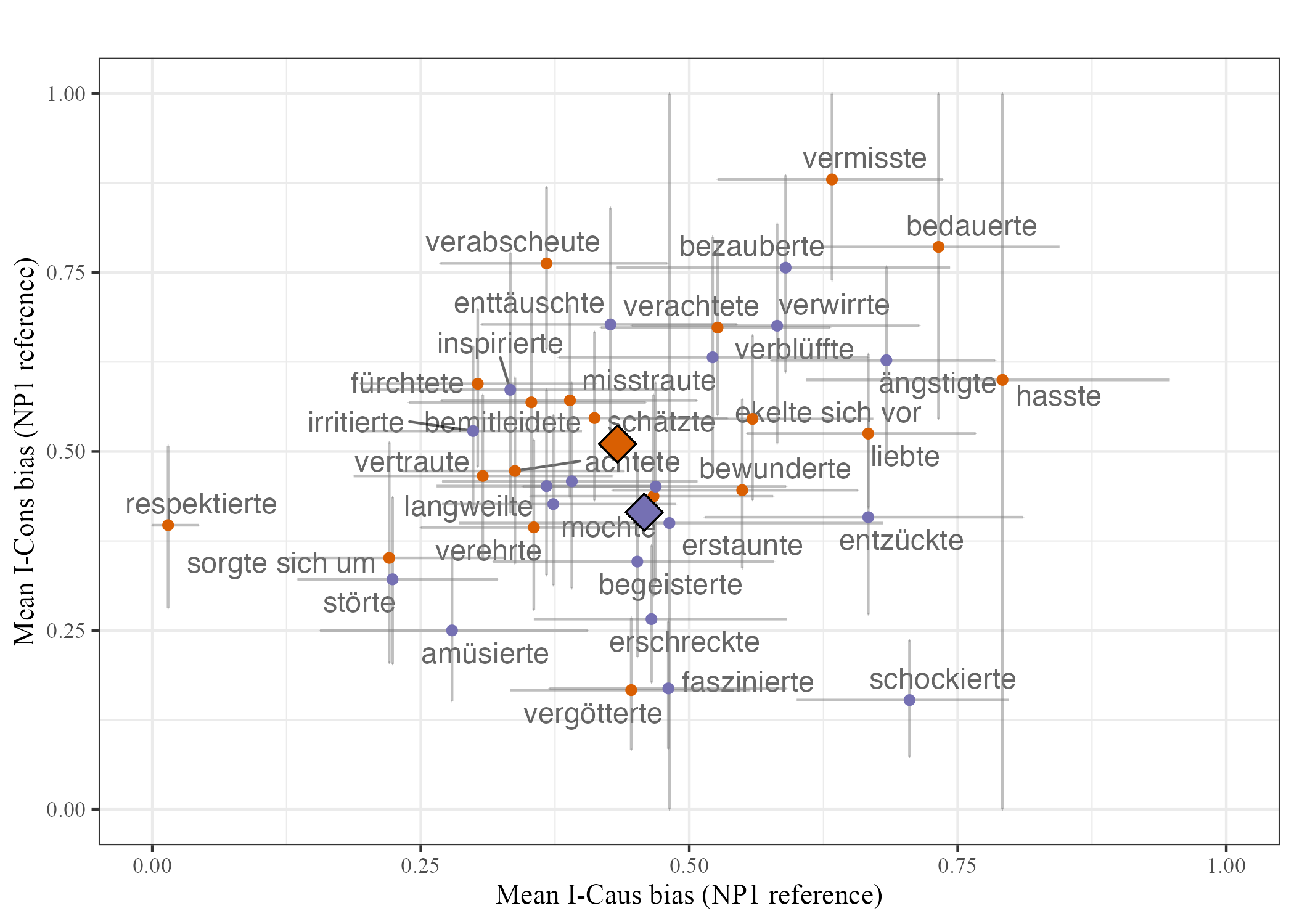}%
\caption{XGLM 0.564 B}\label{coref_fb_0564}
\end{subfigure}
\begin{subfigure}{0.32\textwidth}
\includegraphics[width = \textwidth]{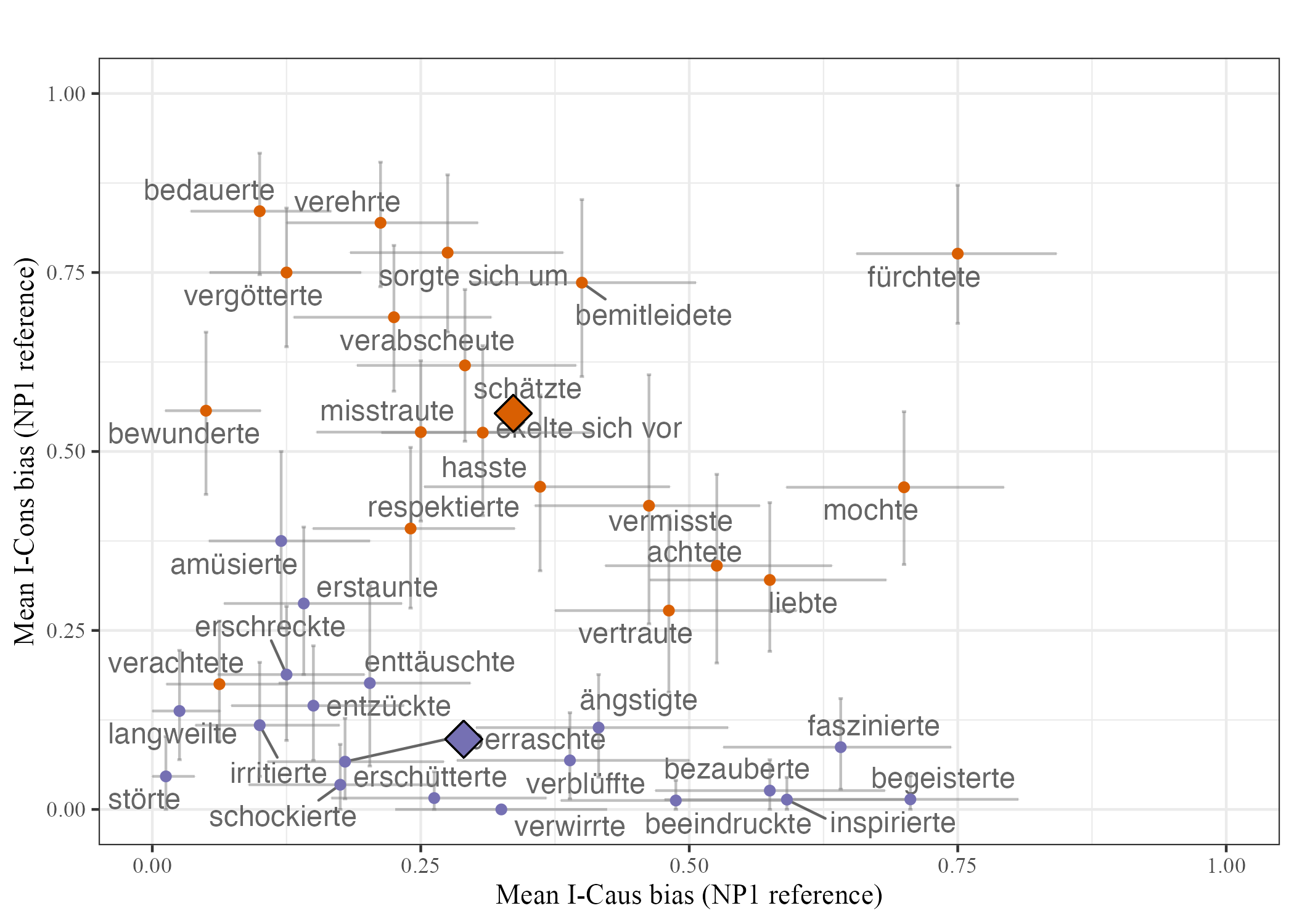}
\caption{XGLM 1.7 B}
\end{subfigure}
\begin{subfigure}{0.32\textwidth}
\includegraphics[width = \textwidth]{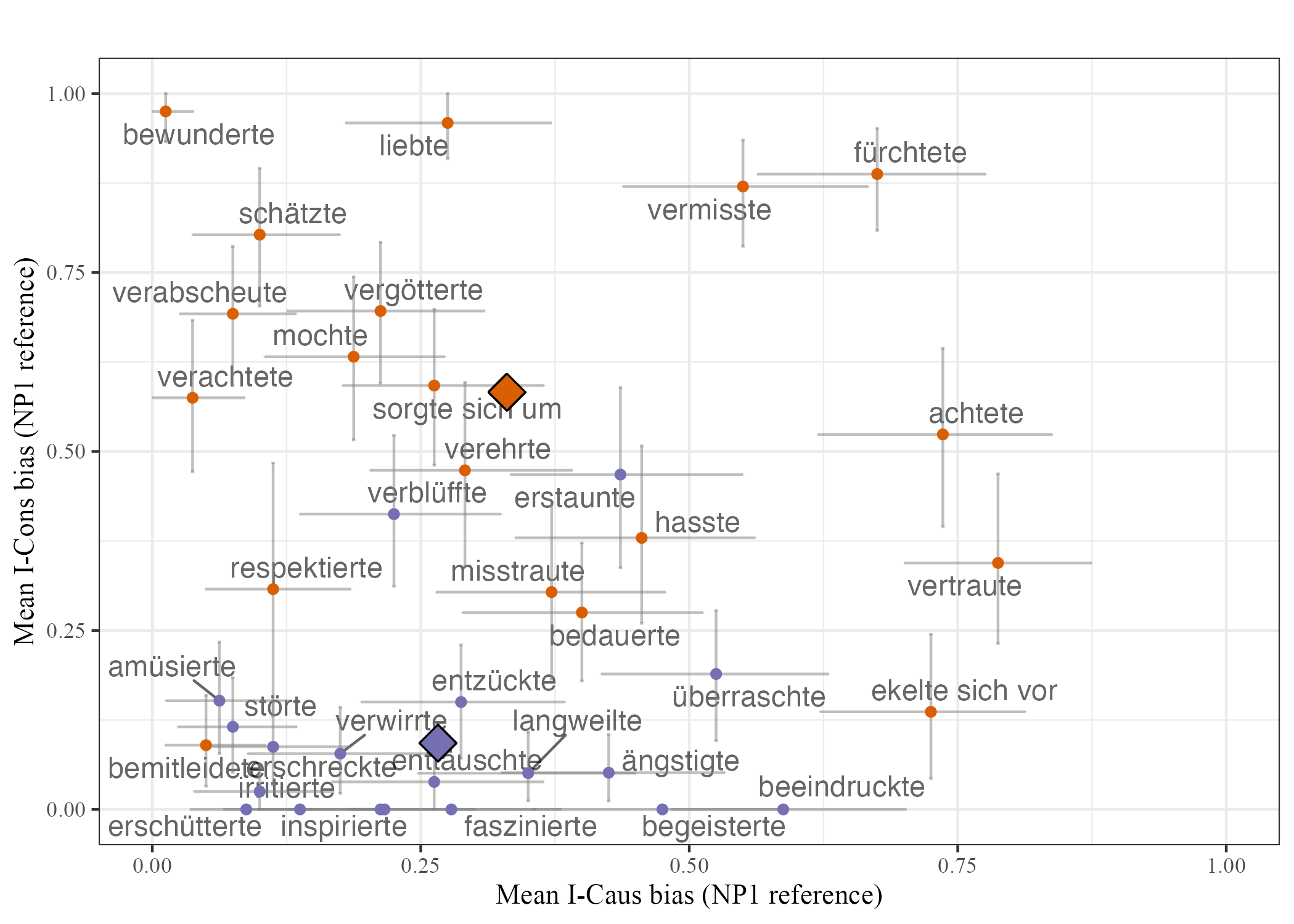}
\caption{XGLM 2.9 B}
\end{subfigure}

\begin{subfigure}{0.32\textwidth}
\includegraphics[width = \textwidth]{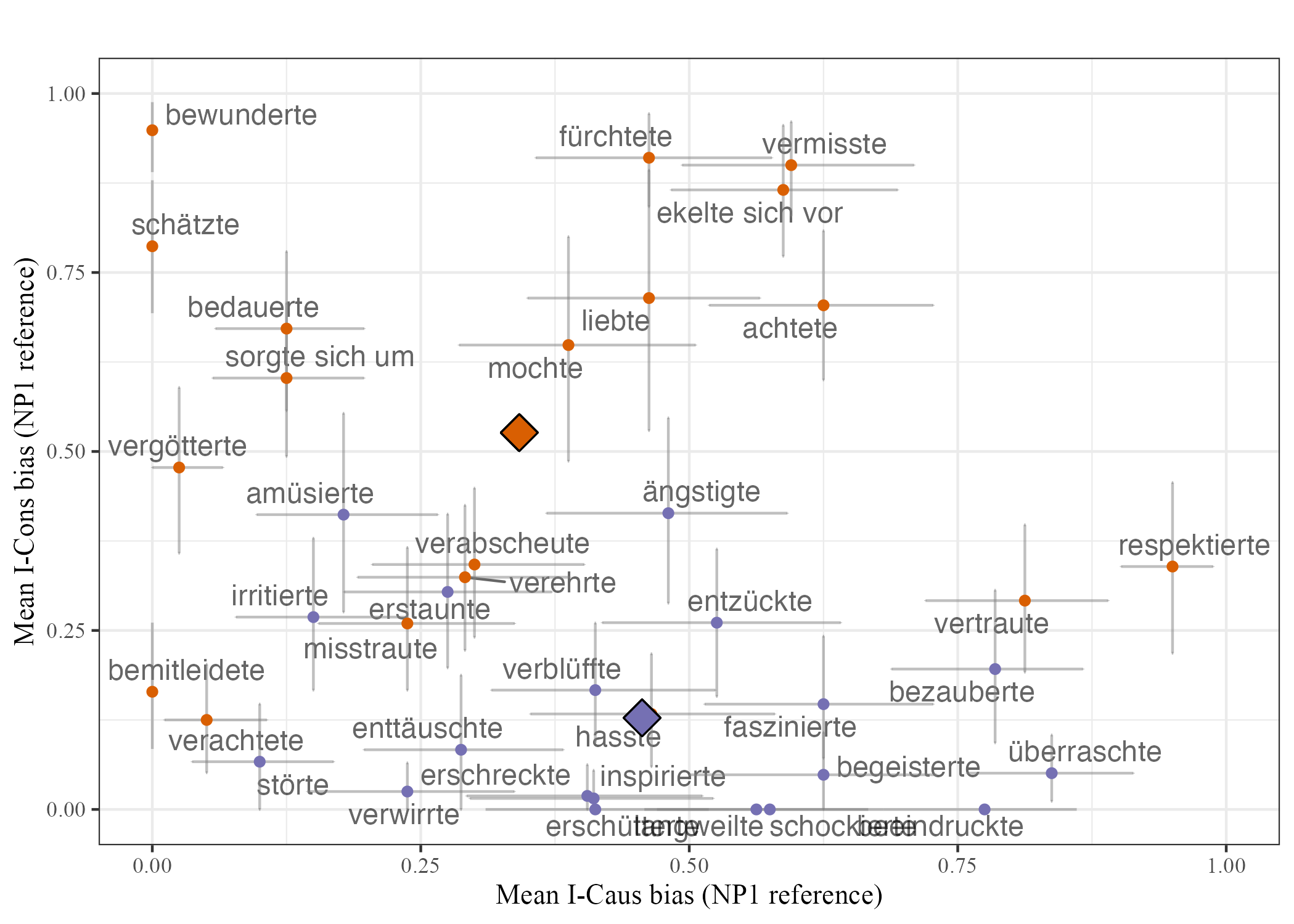}
\caption{XGLM 4.5 B}
\end{subfigure}
\begin{subfigure}{0.32\textwidth}
\includegraphics[width = \textwidth]{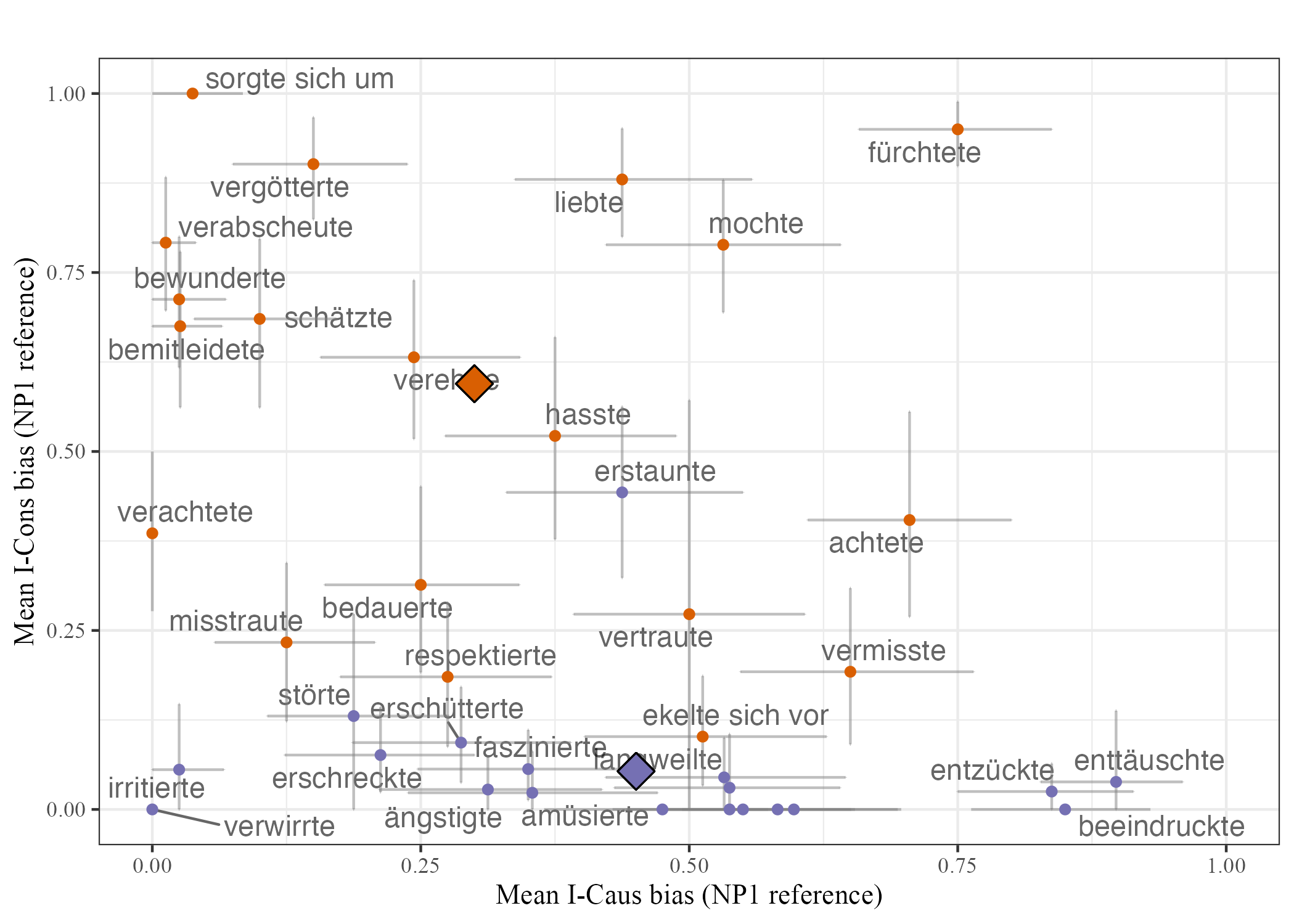}
\caption{XGLM 7.5 B}\label{coref_fb_75}
\end{subfigure}

\begin{subfigure}{0.32\textwidth}
\includegraphics[width = \textwidth]{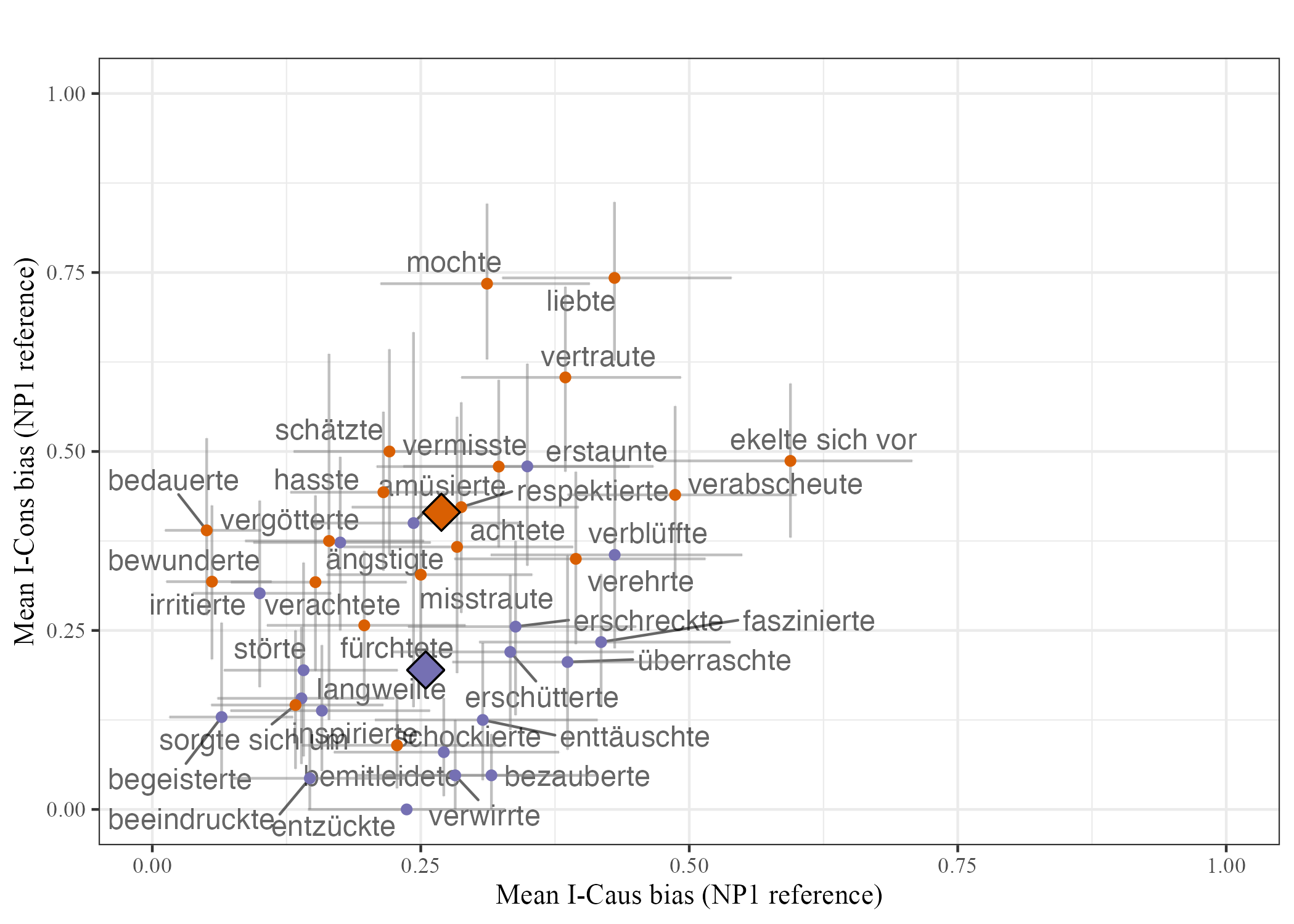}
\caption{German BLOOM 0.350 B}
\end{subfigure}
\begin{subfigure}{0.32\textwidth}
\includegraphics[width = \textwidth]{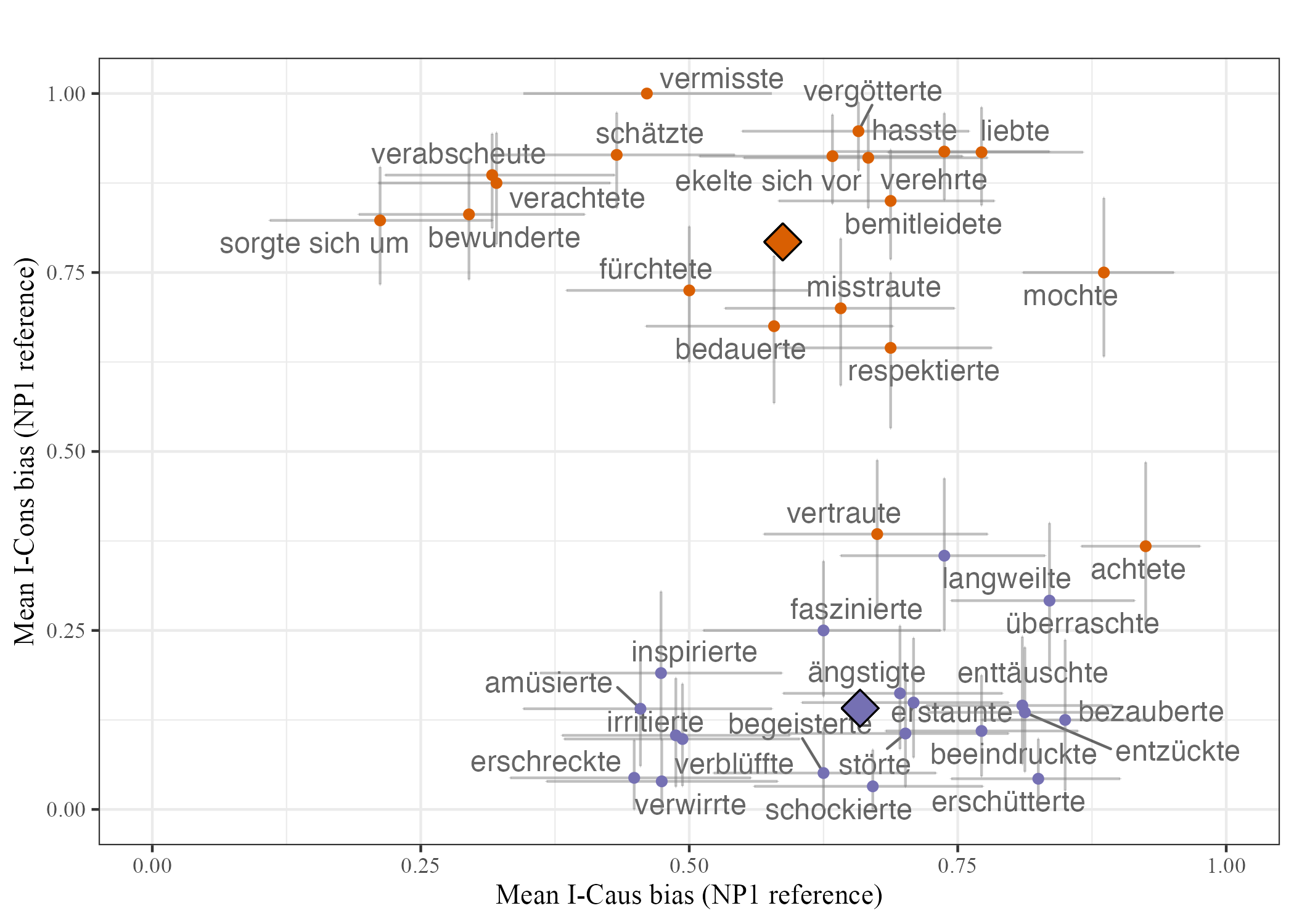}
\caption{German BLOOM 1.5 B}\label{coref_bloom_15}
\end{subfigure}
\hspace{1pt}
\begin{subfigure}{0.32\textwidth}
\fcolorbox{white}{SkyBlue}{\includegraphics[width = \textwidth]{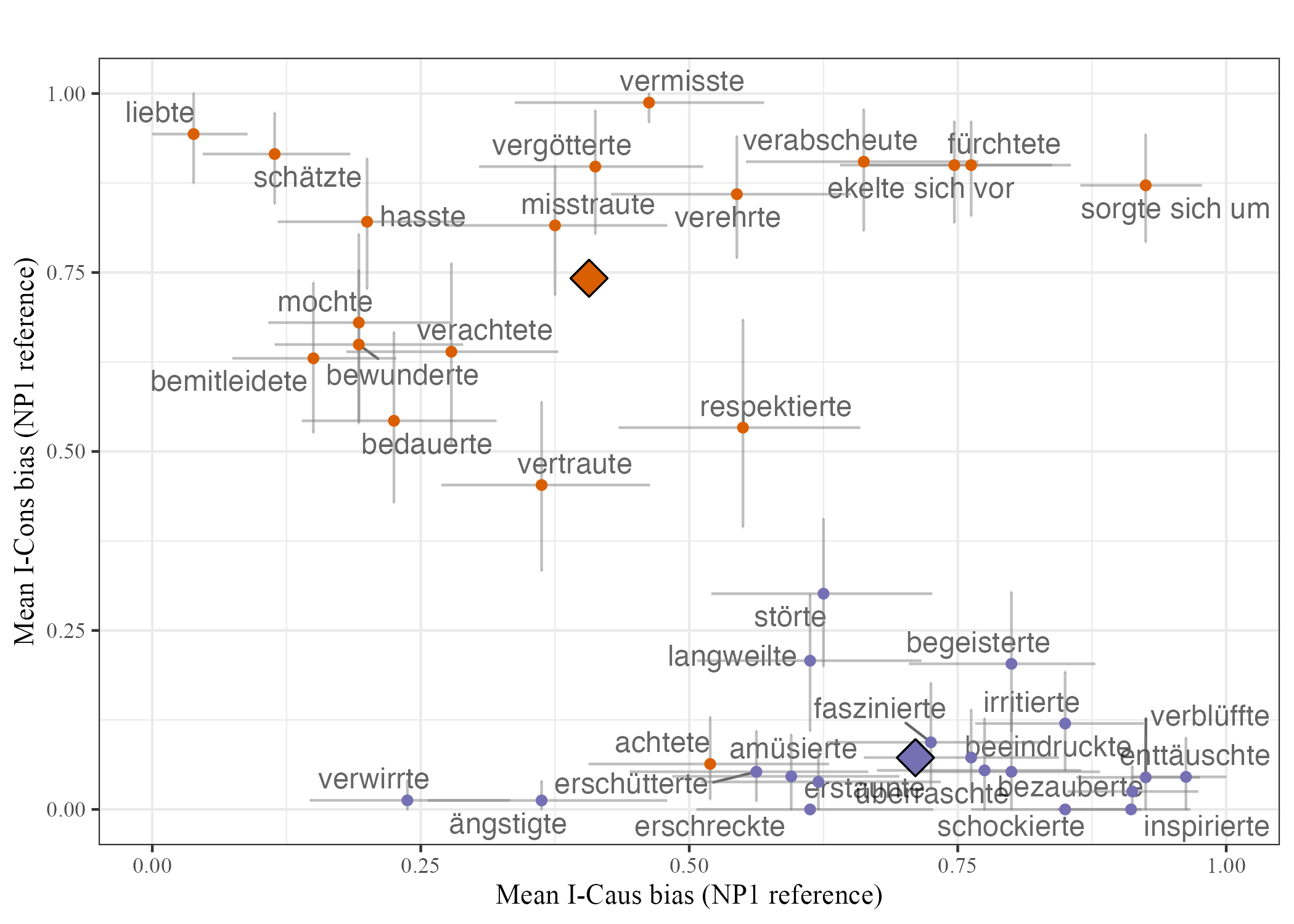}}
\caption{German BLOOM 6.4 B}\label{coref_bloom_64}
\end{subfigure}
\caption{I-Caus and I-Cons biases of individual verbs for human data (Figure \ref{coref_humans}) and the LLMs used in this study. I-Caus biases are plotted on the x axis and I-Cons biases on the y axis (1 = subject coreference, 0 = object coreference). The average I-Caus and I-Cons biases for stimulus-experiencer (stim.-exp.) and experiencer-stimulus (exp.-stim.) verbs are plotted as a diamond-shaped points.}\label{coreference_plots}
\end{figure}

The I-Caus and I-Cons biases for individual verbs are plotted in Figure \ref{coreference_plots} for all LLMs in addition to the human ``gold standard'' data \citep[from][in the light grey frame in the upper left corner]{SolstadBott2022}.  The same automatic annotation was employed for all data. As can be seen from Figures \ref{coref_gpt2} through \ref{coref_bloom_64}, no LLM displays the charateristic picture associated with the strong negative correlation between I-Caus and I-Cons biases for the two verb classes in the human data (see Table \ref{inferential:coreference} for details for individual LLMs). One LLM even makes no significant distinction between the two verb classes whatsoever (Figure \ref{coref_fb_0564}). While the remaining nine LLMs distinguish between I-Cons biases (vertical axis) for the two verb classes, only one LLM displays a significant difference between the I-Caus biases for the two verb classes (horizontal axis), which is why this model is deemed to be the best with regard to this task (Figure \ref{coref_bloom_64}, in the blue frame). Overall, LLMs performance may be said to increase with size (e.g., Figures \ref{coref_fb_75} and \ref{coref_bloom_64}) and performance is also generally better for monolingual LLMs (Figures \ref{coref_bloom_15} and \ref{coref_bloom_64}) than for multilingual ones (e.g, XGLM models) with respect to next-mention coreference bias.

\begin{table}[ht!]
\begin{center}
\setlength\tabcolsep{.2em}
\begin{tabular}{lrrlcrrlcrrlcrrl}\toprule
    Data & \multicolumn{3}{c}{Correlation} & & \multicolumn{3}{c}{\textsc{v.class}$\times$\textsc{bias t.}} & & \multicolumn{3}{c}{\textsc{v.class}} & & \multicolumn{3}{c}{\textsc{v.class}} \\
    & \multicolumn{3}{c}{I-Caus/I-Cons} & & \multicolumn{3}{c}{Interaction} & & \multicolumn{3}{c}{I-Caus} & & \multicolumn{3}{c}{I-Cons} \\
    \cmidrule(lr){2-4} \cmidrule(lr){6-8} \cmidrule(lr){10-12} \cmidrule(lr){14-16}
    & \multicolumn{3}{c}{$r =$} & & \multicolumn{3}{c}{$\chi^2(1) =$} & & \multicolumn{3}{c}{$\chi^2(1) =$}  & & \multicolumn{3}{c}{$\chi^2(1) =$} \\ \midrule
    \rowcolor{lightgray}\textsc{human} & \Checkmark & -.94 & *** & & \Checkmark & 1161.3 & *** & & \Checkmark & 681.3 & *** & & \Checkmark & 487.8 & *** \\
    G-GPT-2 & & $.24$ & & & \Checkmark & 27.9 & *** & & & 3.5 & & & \Checkmark & 6.8 & ** \\
    mGPT & & $.04$ & & & \Checkmark & 22.7 & *** & & & 0.2 & & & \Checkmark & 33.2 & *** \\
    XGLM 0.564B & & $.27$ & & & & 1.7 & & & & NA & & & & NA & \\
    XGLM 1.7B & & $-.08$ & & & \Checkmark & 19.8 & *** & & & 0.3 & & & \Checkmark & 45.8 & *** \\
    XGLM 2.9B & & $.04$ & & & \Checkmark & 16.0 & *** & & & 0.6 & & & \Checkmark & 28.9 & *** \\
    XGLM 4.5B & & $-.19$ & & & \Checkmark & 26.2 & *** & & & 3.8 &  & & \Checkmark & 26.2 & *** \\
    XGLM 7.5B & \Checkmark & $-.33$ & * & & \Checkmark & 17.2 & *** & & & 1.9 & & & \Checkmark & 27.1 & *** \\
    G-Bloom 0.350B & \XSolidBrush & $.36$ & * & & & 13.0 & *** & & & .2 & & &  & 16.4 & *** \\
    G-Bloom 1.5B & & $-.28$ & & & \Checkmark & 46.2 & *** & & & 2.1 & & & \Checkmark & 60.2 & *** \\
    \rowcolor{SkyBlue} G-Bloom 6.4B & \Checkmark & $-.48$ & ** & & \Checkmark & 52.1 & *** & & \Checkmark & 15.9 & *** & & \Checkmark & 49.9 & *** \\ \bottomrule
\end{tabular}
\end{center}
\caption{Statistics for the next-mention coreference bias performance of the different LLMs as compared to the human data (from left to right):\ i) Correlation between I-Caus and I-Cons biases reported as Pearson's $r$, ii) Interaction between the factors \textsc{verb class} and \textsc{bias type} (model comparison with model with two main effects), iii) Main effect of \textsc{verb class} for I-Caus continuations, iv) Main effect of \textsc{verb class} for I-Cons continuations, comparison with intercept only models in the latter two cases. Significance levels indicated by asterisks:\ * = $p < .05$; ** = $p < .01$; *** $= p < .001$. LLM values marked with a \Checkmark{} signify a result in the direction of the human data, while a \XSolidBrush{} indicates a result contrary to the human data.}\label{inferential:coreference}
\end{table}%

Although statistical analysis shows that the LLMs can succesfully distinguish between stimulus-experiencer and experiencer-stimulus verbs with respect to I-Cons bias, there are important qualitative differences between the performance of LLMs and humans. Thus, it may be noted that all LLMs display a stronger overall object bias as compared to the human data, being skewed towards the lower left corner. This observation is in line with the findings by \citet{kementchedjhieva2021john} and \citet{zarriess2022isn} for language model IC bias in general. Thus, while stimulus-experiencer verbs display a strong object I-Cons bias -- as expected -- in the range 80.5--94.7\%  (humans:\ 95.2\%), only two LLMs show an I-Cons subject bias above 60\% for experiencer-stimulus verbs (humans:\ 77.9\% subject bias). This can be seen as a reflection of a preference for recency in coreferential depedencies \citep[see also the general discussion]{GernsbacherHargreaves1988}.

As already mentioned, only one model (Figure \ref{coref_bloom_64}) displayed a statistically significant difference between the I-Caus biases for stimulus-experiencer and experiencer-stimulus verbs, with the difference amounting to 30.4\% for the two verb classes (compared to a difference of 77.9\% in the human gold standard). Even here, however, the I-Caus bias of the two verb classes isn't very close to the human basis of comparison. In particular, the experiencer-stimulus verbs are not as strongly object-biased in the LLM (59.3\% object bias) as compared to humans (94.7\% object bias). Thus, in addition to the LLMs' general tendency towards object bias, there seems to be a difference between the two bias types. In general it may seem surprising that performance is better for I-Cons than for I-Caus, given that the I-Caus-associated connective, \textit{weil} `because', is more frequent in corpus data -- meaning that the models will have encountered it more often during training. One reason may be that in general, explanations display more variation than consequences with regard to the syntactic position (they may occur in cause-effect and effect-cause order) and the types of relation they may enter (see, in particular \citeauthor{SolstadBott2022} \citeyear{SolstadBott2022}:\ Exp.\ 4; see also \citeauthor{Sweetser1990} \citeyear{Sweetser1990} for a more general discussion), possibly making them harder to sample.

In experiments involving human participants, \textbf{\textsc{gender order}} is usually included merely as a counterbalancing factor (comparing, e.g., \textit{Anna fascinated Peter because \dots} and \textit{Peter fascinated Anna because}), since mostly no evidence as to an influence of the gender of arguments is found \citep[see, however,][]{Ferstletal2011}. For the LLMs, we nevertheless chose to include this factor as a main factor in our statistical analysis (see above), as earlier studies have indicated that the gender of antecedents may influence the bias \citep{zarriess2022isn}. Indeed, when comparing regression models including \textsc{gender order} as a main factor with models without this factor, \textsc{gender order} turned out to significantly affect coreference biases in all but two LLMs (XGLM 2.9B and German Bloom 6.4B), and also in the human data. These effects weren't investigated further in regression models, but the following observations can be made based on descriptive statistics:\ In the human data, there is a (weak) tendency to establish coreference to male subject arguments, whereas in almost all LLMs, coreference is more frequently established to female arguments. Thus, in the domain of IC the male gender bias often discussed in relation to language models \citep[see][]{stanovsky-etal-2019-evaluating, gautam2024, nemani-2024} does not seem to manifest itself.

\section{Experiment 2:\ Coherence Bias}
\label{sec:exp2}

Next, we turn to the coherence bias, which concerns what types of coherence relations are triggered following IC verbs. Again, the gold standard data will consitute our point of departure.

Since the seminal study by \citet{Kehleretal2008}, IC verbs have repeatedly been found to trigger significantly more explanations than other discourse relations after a full stop in prompts like the following:

\ex. \a. Mary fascinated Peter. \dots{} \HandPencilLeft
\b. Mary admired Peter. \dots{} \HandPencilLeft

Thus, \citet{SolstadBott2022} found that stimulus-experiencer and experiencer-stimulus verbs evoked around 60\% explanations for both verb classes (stimulus-experiencer:\ 58.2\%; experiencer-stimulus:\ 60.2\%), fitting the overall results from \citet{Kehleretal2008}, who did not control for verb class. Importantly, this is a bias going far beyond the overall preference for causal relations argued for in relation to the  ``causality-by-default'' \citep[see][and much subsequent work]{Sanders2005}.

In assessing the coherence relation, human annotators typically test for the suitability of a particular relation by inserting unambiguous connectives like \textit{because} for explanations or \textit{and so} for consequence/result relations \citep{Rohde2008,Kehleretal2008,BottSolstad2014,BottSolstad2021,SolstadBott2022,BottSchrumpfMichaelisSolstad2023}. However, as previous research  shows that most of the relations after a full stop are not explicitly marked with a discourse connective \citep{SolstadBott2022}, it is difficult to reliably elicit the discourse relation using automatic annotation. In particular with respect to causal relations, it has been shown that state of the art transformer language models make surprisingly many errors with inferring the appropiate discourse connective \citep{chang2024language, pandia2021sorting}, which is similar to the task also employed in manual annotation described above.\footnote{Performance still is not sufficient for our purposes in automated discourse labellers trained with multiple labels and label distributions \citep{yung2022label}, which are not at hand for Implicit Causality data often forcing decisions between labels.} For the present evaluation of coherence biases in LLMs we therefore decided to use prompts that would be more prone to triggering an explicit marking of discourse relations than full stop prompts. 
To this end, we prompted continuations after a comma instead of a full stop. Since such an experiment has not, at least to our knowledge, been conducted before, we present the data in some detail here.

40 items were constructed from those in Experiment 1 by removing the connective, which resulted in a 2 ($\times$ 2) design investigating the factor \textsc{verb class} (\textit{stimulus-experiencer} vs.\ \textit{stimulus-experiencer}) including the counter-balancing factor \textsc{gender order} (\textit{female subject-male object} vs.\ \textit{male subject-female object}):%
\footnote{The stimulus-experiencer verb \textit{gefallen} `please' was replaced by \textit{belustigen} `amuse', which displayed a strong subject bias in \citet{BottSchrumpfMichaelisSolstad2023}.}

\ex.\label{ex:coh-design1} \textbf{Stimulus-experiencer verb \textit{faszinieren} `fascinate'}
\a. \textit{Maria faszinierte Peter, \dots} \HandPencilLeft \hfill (fem.-masc.)\\
`Maria fascinated Peter, \dots'
\b. \textit{Peter faszinierte Maria, \dots} \HandPencilLeft \hfill (masc.-fem.)\\
`Peter fascinated Mary, \dots'

\ex.\label{ex:coh-design2} \textbf{Experiencer-stimulus verb \textit{bewundern} `admire'}
\a. \textit{Emma bewunderte Karl, \dots} \HandPencilLeft \hfill (fem.-masc.)\\
`Emma admired Karl, \dots'
\b. \textit{Karl admired Emma, \dots} \HandPencilLeft \hfill (masc.-fem.)\\
`Karl admired Emma, \dots'

30 native German speakers (9 female, 21 male; mean age 24.9 years, range 18-34 years) were recruited via Prolific (payment:\ GBP 5.50). They received each verb in one of the two gender-orders, using 40 male and 40 female names. The experiment was implemented with PCIbex \citep{Zehr} and participants received the experimental trials in a single block in individually randomized order.

The data were annotated for discourse relations using insertion tests as in \citet[Exp. 2]{SolstadBott2022}. The first author applied tests for \textsc{explanation} (\textit{because} insertion test), \textsc{consequence} (\textit{sodass} test), \textsc{contrast} (\textit{aber} test), \textsc{elaboration} (\textit{und zwar} `namely' test), \textsc{occasion} (\textit{afterwards} test), \textsc{background} (\textit{während} `while' test), as well as \textsc{ambiguous} (multiple insertions equally plausible) and \textsc{other} (none of the insertion tests yields a plausible interpretation). In addition, all explicitly marked discourse connectives were coded according to the list in \citet{Breindletal2014} by which we hoped to include a broader range of explanatory relations.%
\footnote{The explanatory connectives included this way were:\ \textit{denn} and \textit{da} `since', \textit{nachdem} `after' and \textit{indem} `by'. While the latter two are in principle ambigous, they turned out to be used almost exclusively in an explanatory sense in the current experiment.} %
Finally, the data were also annotated with respect to general sensicality, OVS interpretations of the prompt and for coreference of anaphoric expressions applying the same annotation as in Experiment 1. The discourse relation annotation was checked for inter-annotator agreement with a second linguistically trained annotator on a random sample of 200 of the 1,200 continuations (Cohen's $\kappa = .75$). 

\begin{figure}[t]
\begin{center}
\includegraphics{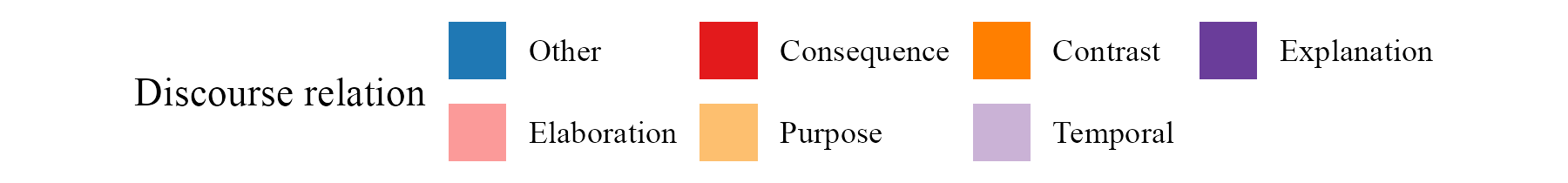}
\includegraphics[width=0.6\textwidth]{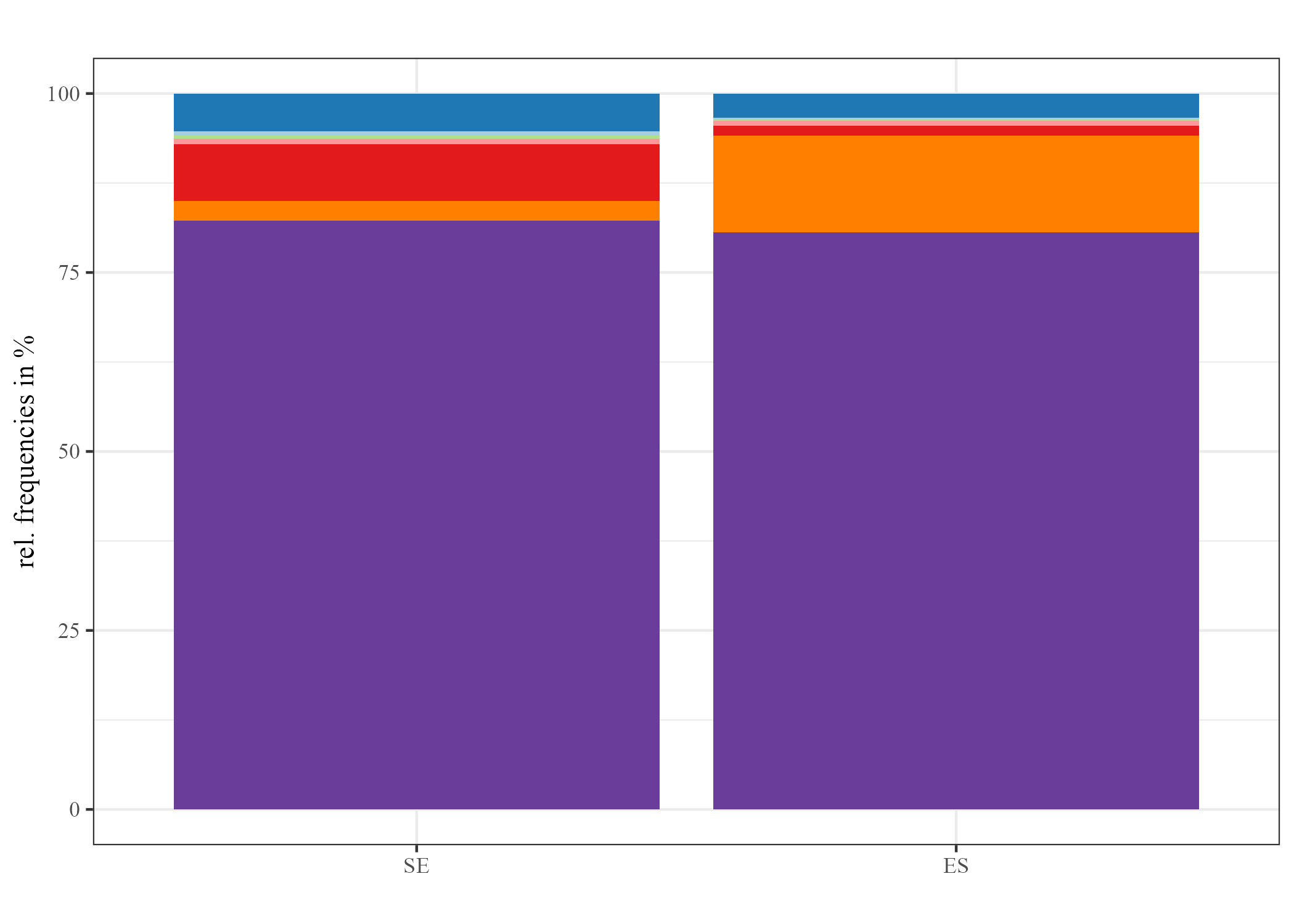}
\caption{Distributions of discourse relations produced in continuations after prompts ending in a comma with stimulus-experiencer (SE) and experiencer-stimulus (ES) IC verbs.}\label{Figure_DR_human}
\end{center}
\end{figure}

In the data, we identified a total of 27 discourse connectives, among which 15 occurred clause-initially. In line with our expectations, most \textsc{explanations} provided for the comma prompts were explicitly marked by means of discourse connectives (677 out of 859 \textsc{explanations} in total). The distribution of discourse relations is shown in Figure \ref{Figure_DR_human}.  Continuations after a comma gave rise to more explanations than observed for continuations after a full stop in \citet{SolstadBott2022}. Comma prompts with stimulus-experiencer verbs received 82.2\%, while prompts with experiencer-stimulus verbs received 80.6\% explanations, respectively \citep[compared to 58.2\% for stimulus-experiencer and 60.2\% for experiencer-stimulus verbs in][]{SolstadBott2022}. Thus, the comma-prompts gave rise to an even stronger coherence bias than the one observed for continuations after a full stop. On the one hand, this may be related to the fact that we included a wider range of connectives, although \textit{weil} `because' continuations only made up 35.9\% of those explanations. Another reasons lies in the difference in punctuation. A comma should trigger more embeddings, and since explanations are subordinating discourse relations often, but not necessarily, associated with syntactic embedding, a greater proportion of explanations (and other embedding relations) is to be expected \citep{KaragjosovaSolstadTA}.

Ignoring the counter-balancing factor of \textsc{gender order} (which was not found to be significant; see the output of the RMD script for details), we conducted a logit mixed-effects regression analysis modeling the likelihood to produce an explanation as a function of the centered fixed of \textsc{verb type} as well as by-participants and by-items random intercepts and slopes of \textsc{verb type}. As in \citet{SolstadBott2022}, we confirmed that explanations made up more than all other relations in sum by investigating the intercept, which differed signficantly from zero with a positive estimate ($\beta = 2.03, \mathit{SE} = .28, z = 7.29$). Further in line with the results in \citet{SolstadBott2022}, we found no significant effect of \textsc{verb class} when comparing this model with a model including only the intercept ($\chi^2(1) = .05, p = .83$).

\subsection{Methods}

Following the design for human participants in \citet[Exp.\ 2]{SolstadBott2022}, LLMs were prompted to provide continuations for 20 stimulus-experiencer and 20 experiencer-stimulus verb prompts as illustrated in \ref{ex:coh-design1} and \ref{ex:coh-design2}. 40 female and 40 male first names were paired such that all pairs were tested in both conditions resulting in a total of 3,040 continuations.

\subsection{Data annotation and selection}

In a first step, we filtered out continuations that could not be assigned a syntactic parse. For annotating the discourse relations, we used DiMLex \citep{stede1998dimlex} to map the explicit discourse connectors to their respective discourse relation. We only included main or subordinate clauses with a clause-initial connective in this process.%
\footnote{In fact, differing from humans, the LLMs produced no clause-internal connectives.} 

Comparing the automatic and manual annotation of the data in experiment 2 for the binary label explanation vs. non-explanation yields an inter-annotator agreement of $\kappa = 0.68$. However, the data contain a lot of paratactical main-clause continuations without an explicit discourse connective, which our automatic annotator ignores, as reliable implicit discourse classification is still an unsolved problem \citep[but see][for a thorough attempt at implementing a German discourse parser]{bourgonje2021shallow}. Incidentally, the language models rarely produced these kinds of continuations. Excluding any such main-clause continuations yields an acceptable agreement of $\kappa = .86$, which more accurately reflects on the performance of the annotator.

Our filtering process led to the exclusion of between 39.9\% and 77.1\% of all data. These relatively large proportions came about since the LLMs produced a large amount of relative clauses and main clauses without a discourse marker (58.1\% and 6.3\%, respectively, in the model with the largest proportion of excluded data points).%
\footnote{Based on the continuations produced by humans, we had intended to include a subset of relative clauses that were explanatory in nature (\textit{Peter admired Mary, who was very good at chess}), but upon manual inspection of the data, it turned out that LLMs did not reliably produce relative clauses of this type (but rather clauses like \textit{Lisa admired Steven, who stayed close to her house}, which arguably does not provide an explanation of the admiration).} %

Although the discourse relations in \cite{stede1998dimlex} are coded according to the taxonomy used for the \textit{Penn Discourse Treebank} in \citet{Webberetal2019}, they were recoded with labels used in \citet{SolstadBott2022} for the purposes of presentation.

\subsection{Statistical analysis}\label{sec:stats_exp2}

For the continuations included in the statistical analysis, we again fitted mixed-effects binomial logistic regression models \citep{Batesetal2015} for each LLM. This time, the binomial dependent variable coded whether the continuation was an explanation or not. As the basic finding in published studies on the human IC coherence bias is a majority of explanations after IC verb prompts, which is not modulated by \textsc{verb class} \citep[a.o.]{SolstadBott2022}, we constructed our analysis to test for explanation being the default coherence relation (analyzing the fixed-effect intercept in a logistic mixed-effects regression analysis) as well as the lack of an effect of \textsc{verb class}. The model therefore included the fixed effects of \textsc{verb class} and (the counter-balancing factor) \textsc{gender order} in addition to their interaction. The random effects structure included by-verb random slopes for \textsc{gender order} in addition to a random intercept for verbs. Beyond this, we proceded as described in section \ref{sec:analysis:coref} with respect to model comparison. In order to investigate the (lack of) effect of \textsc{verb class} we first compared a model including this fixed effect as the only predictor with an ``intercept only'' model without this factor. If \textsc{verb class} turned out to contribute significantly to the proportion of explanations, we investigated the fixed-effects intercepts separately for stimulus-experiencer and experiencer-stimulus verbs, subsetting the data accordingly.

If verb classes did not differ from each other, we directly assessed whether the intercept reliably exceeded zero for any of the two verb types. To check for \textit{explanations as default}, we always (in subset analyses or in the global analysis) compared models with a fixed-effect intercept with models without any fixed effects in one-tailed comparisons testing whether the continuations in fact exhibited reliably more explanations than non-explanations.

\subsection{Results and Discussion}

\begin{figure}[ht!]
\begin{center}
\includegraphics{graphs/coherence/coherence_legend.png}
\end{center}
\begin{subfigure}{0.32\textwidth}
\fcolorbox{white}{black}{
\includegraphics[width = .96\textwidth]{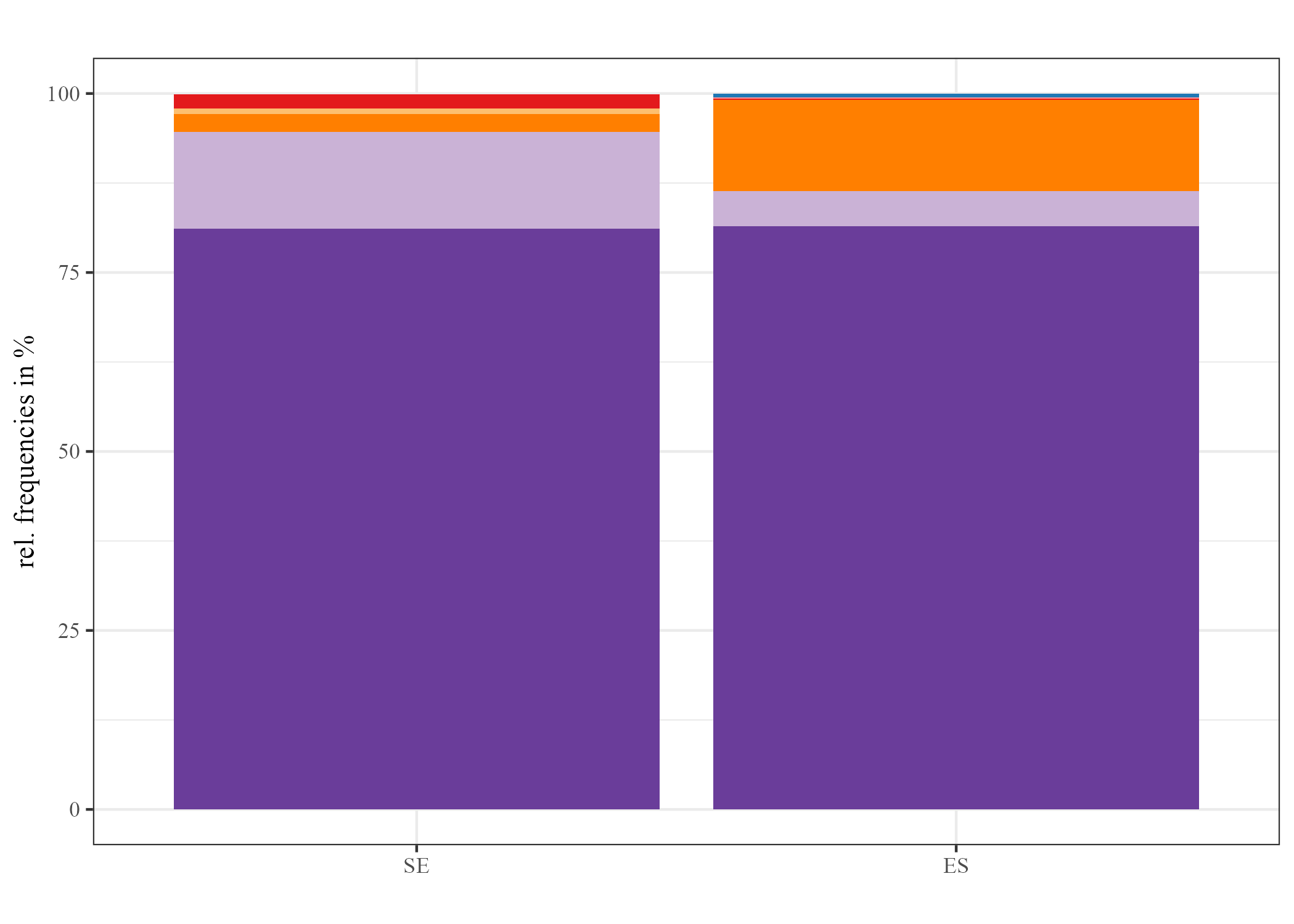}}
\caption{Human data (machine coded)}\label{coherence_humans}
\end{subfigure}
\hspace{0pt}
\begin{subfigure}{0.32\textwidth}
\includegraphics[width = \textwidth]{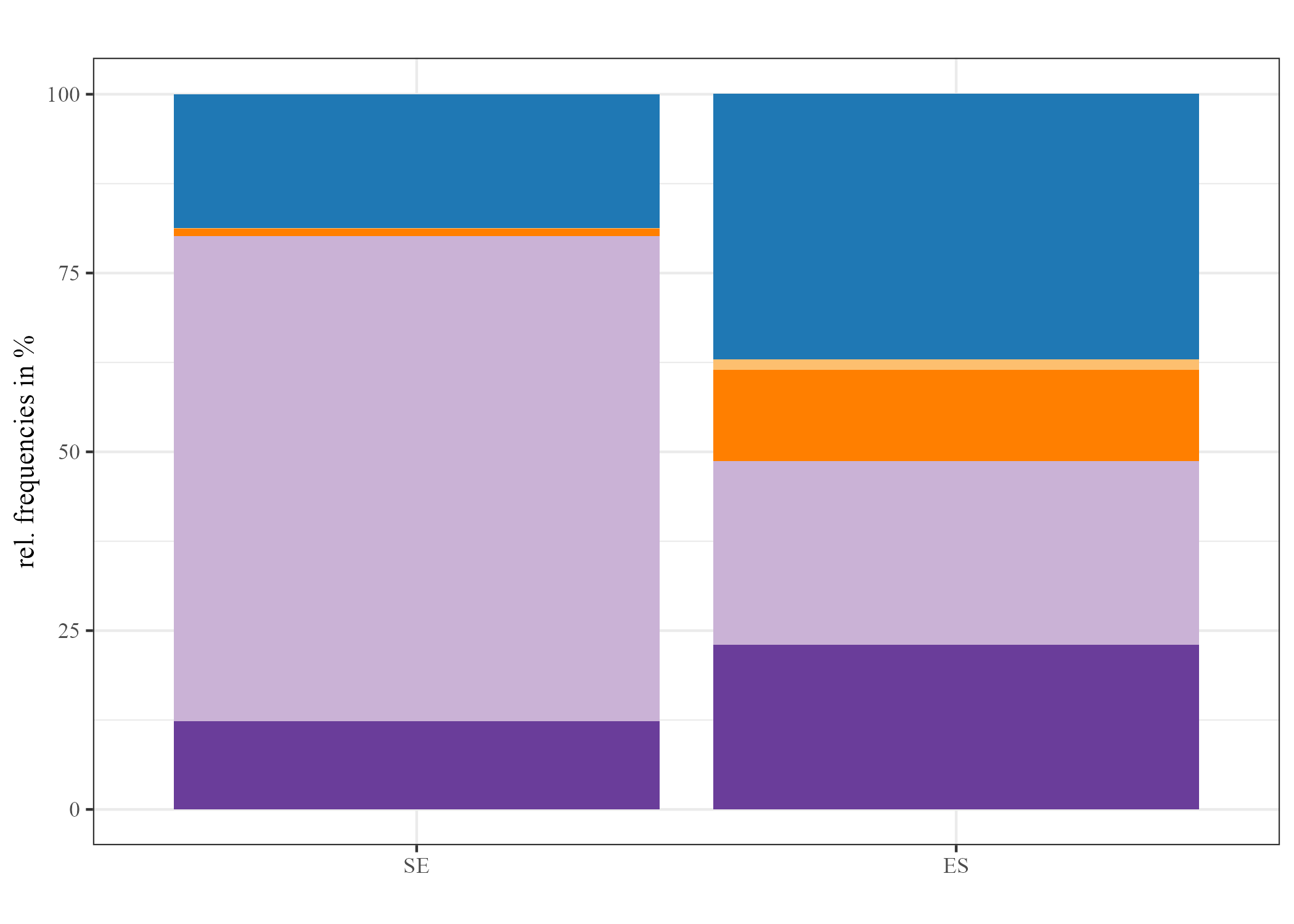}
\caption{German GPT-2}\label{coherence_gpt2}
\end{subfigure}
\begin{subfigure}{0.32\textwidth}
\includegraphics[width = \textwidth]{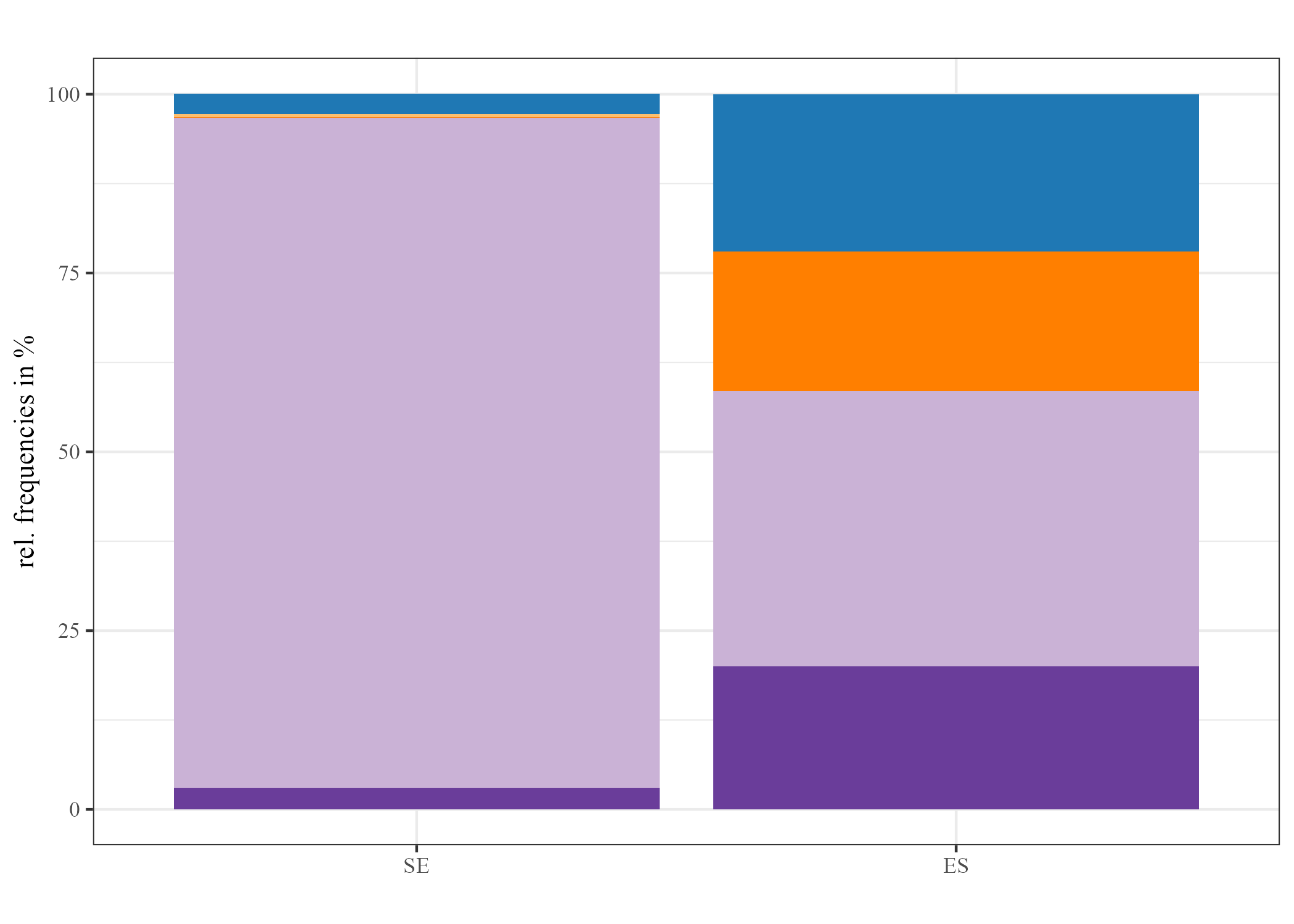}
\caption{mGPT}\label{coherence_mgpt}
\end{subfigure}

\begin{subfigure}{0.32\textwidth}
\includegraphics[width = \textwidth]{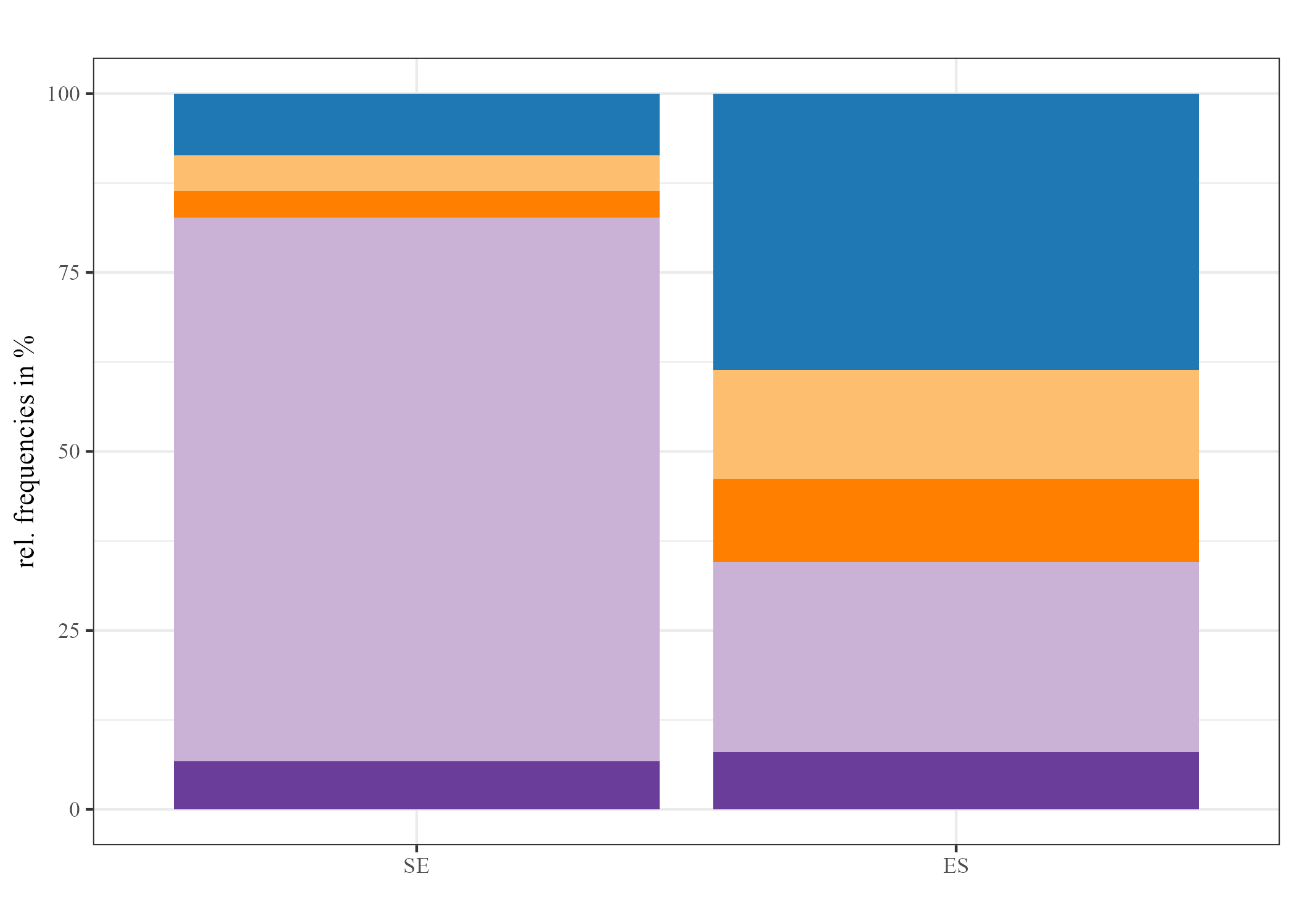}%
\caption{XGLM 0.564 B}\label{coherence_fb_0564}
\end{subfigure}
\begin{subfigure}{0.32\textwidth}
\includegraphics[width = \textwidth]{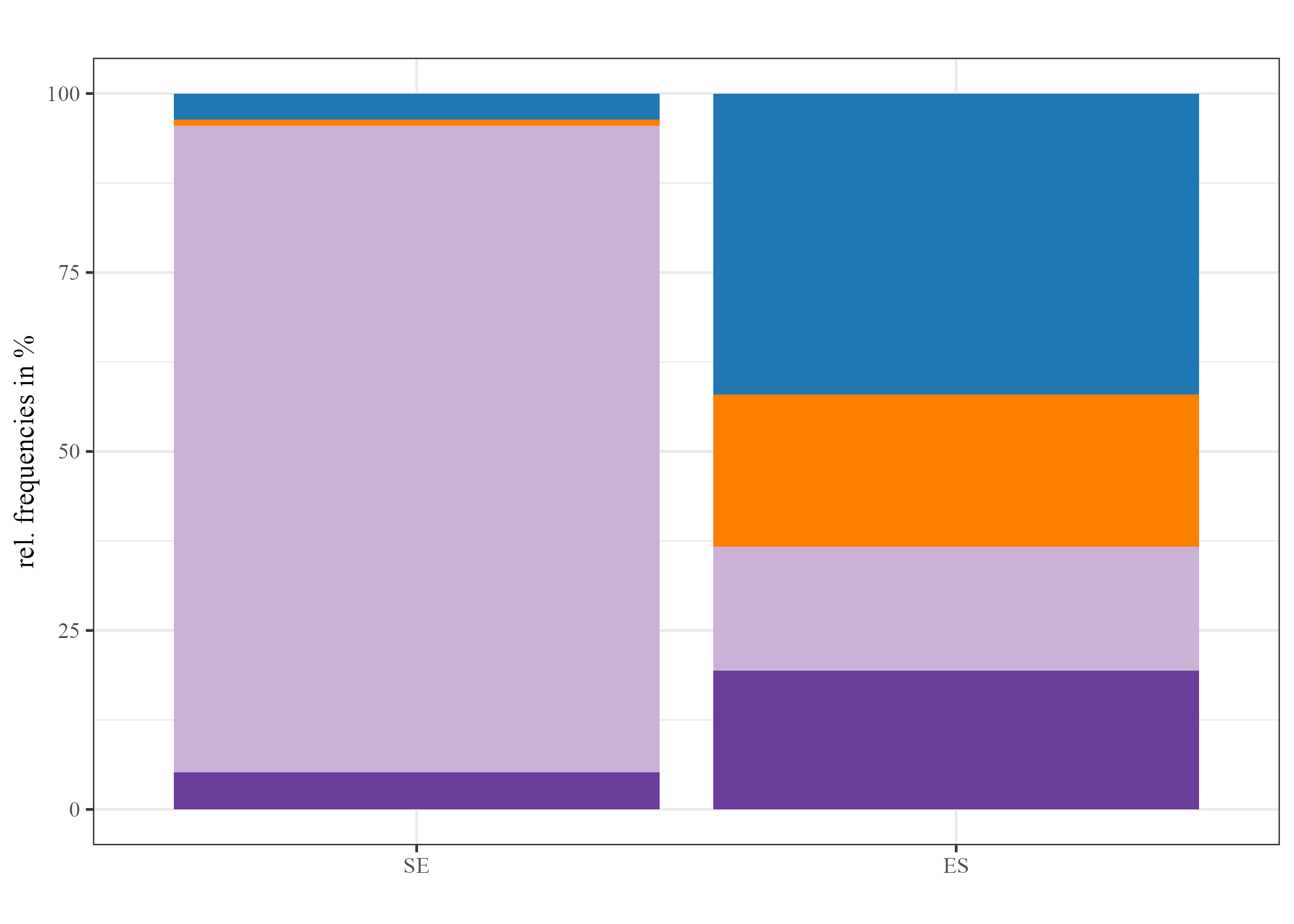}
\caption{XGLM 1.7 B}
\end{subfigure}
\begin{subfigure}{0.32\textwidth}
\includegraphics[width = \textwidth]{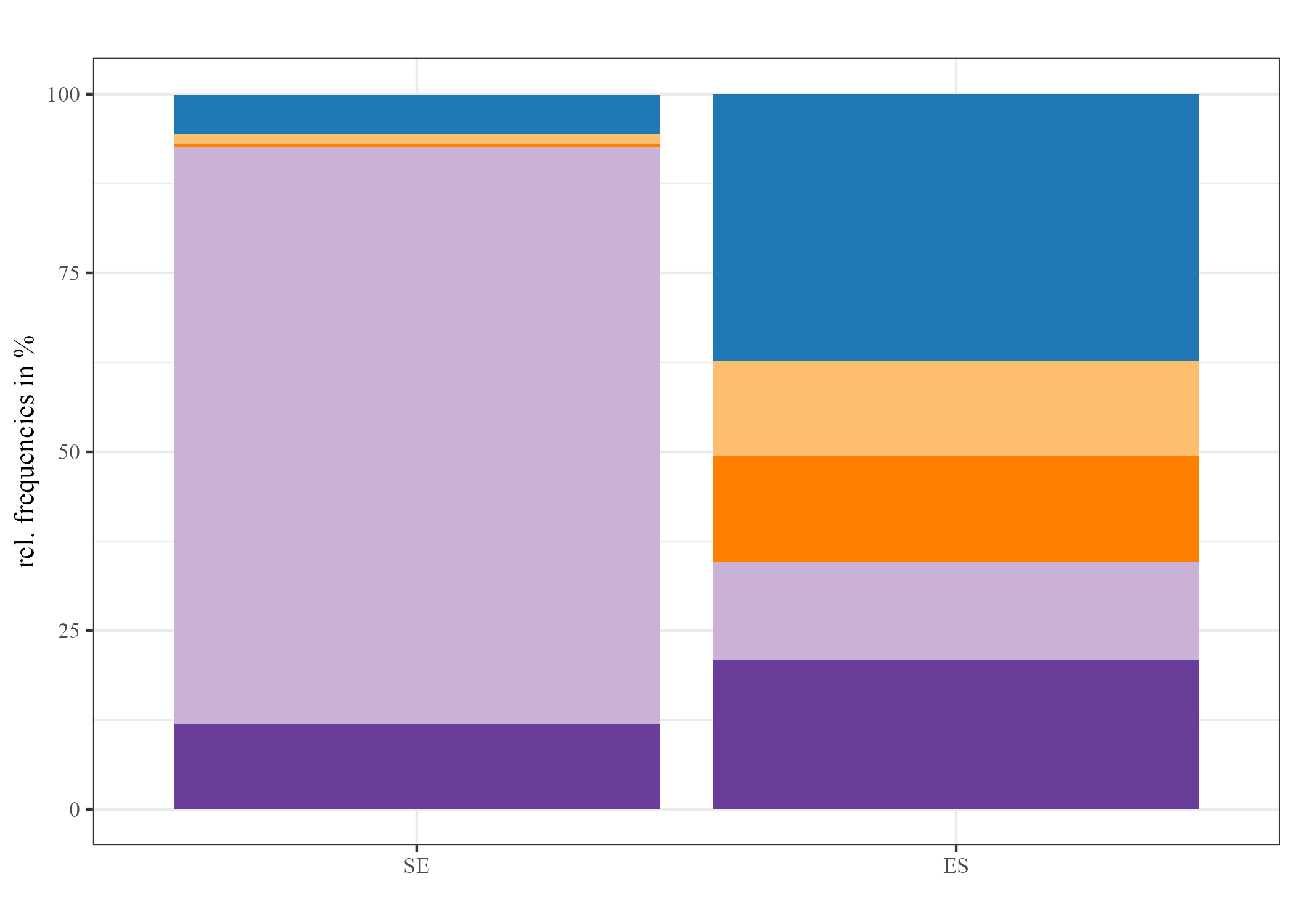}
\caption{XGLM 2.9 B}
\end{subfigure}

\begin{subfigure}{0.32\textwidth}
\includegraphics[width = \textwidth]{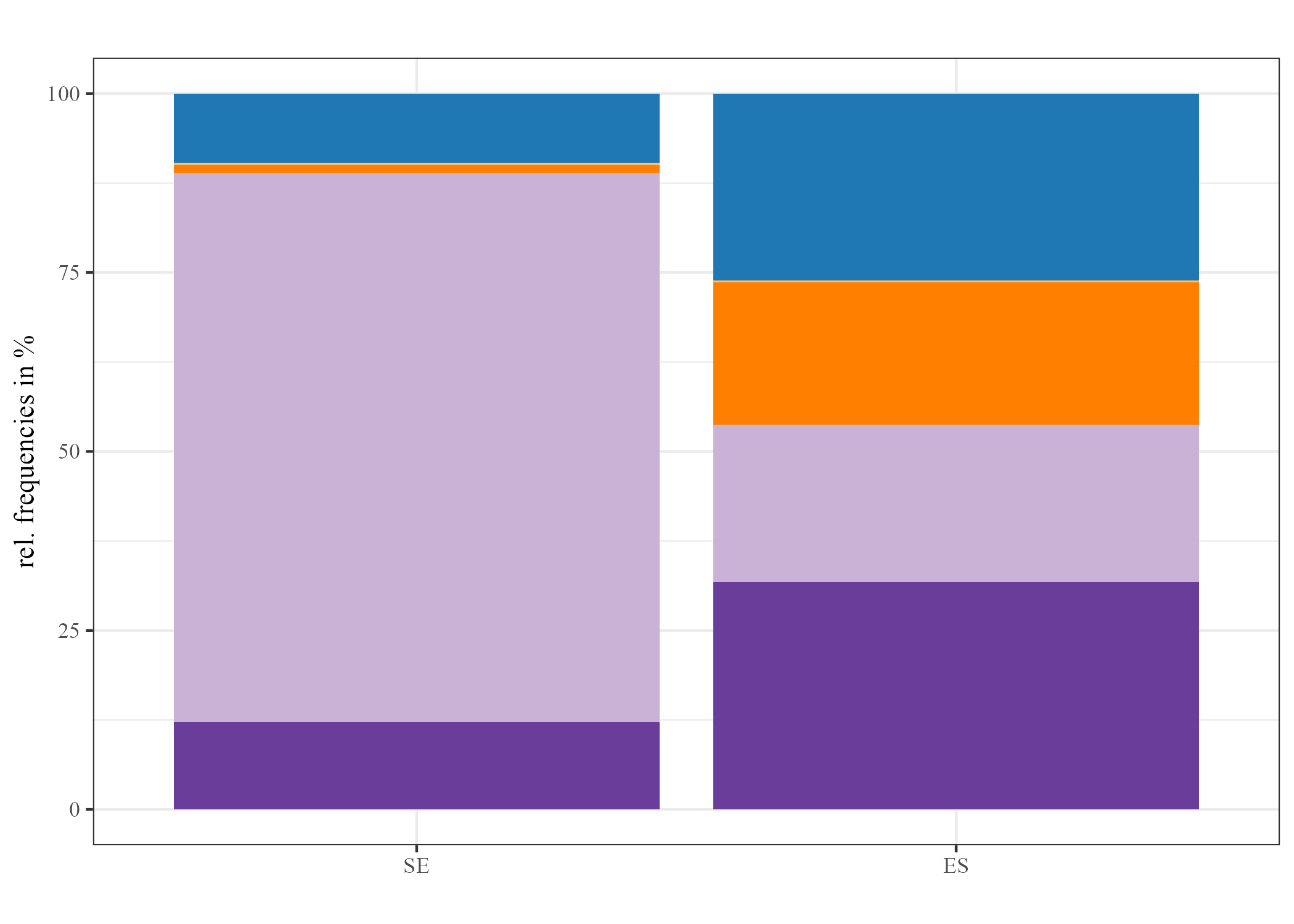}
\caption{XGLM 4.5 B}\label{coherence_fb_45}
\end{subfigure}
\begin{subfigure}{0.32\textwidth}
\includegraphics[width = \textwidth]{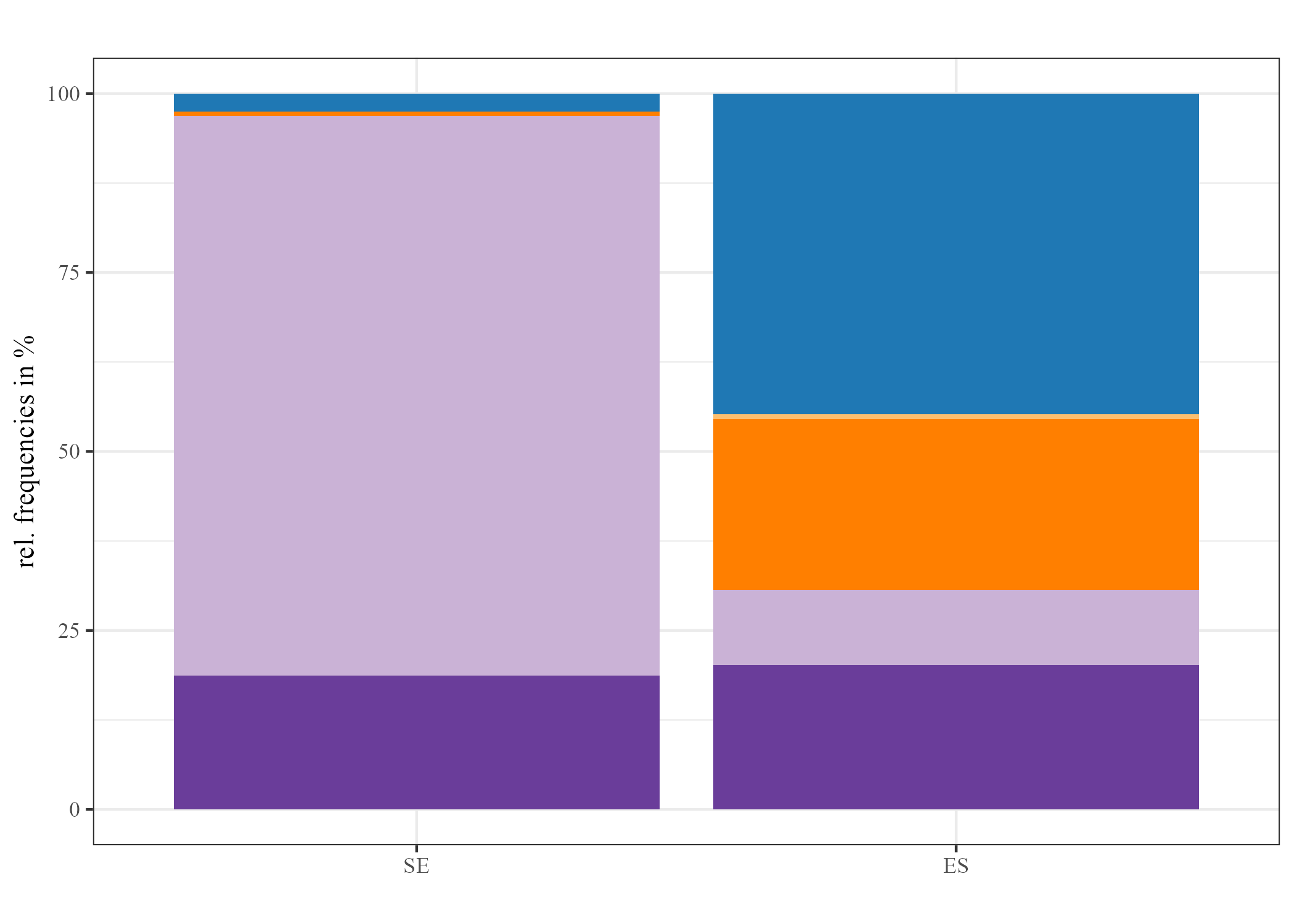}
\caption{XGLM 7.5 B}\label{coherence_fb_75}
\end{subfigure}

\begin{subfigure}{0.32\textwidth}
\includegraphics[width = \textwidth]{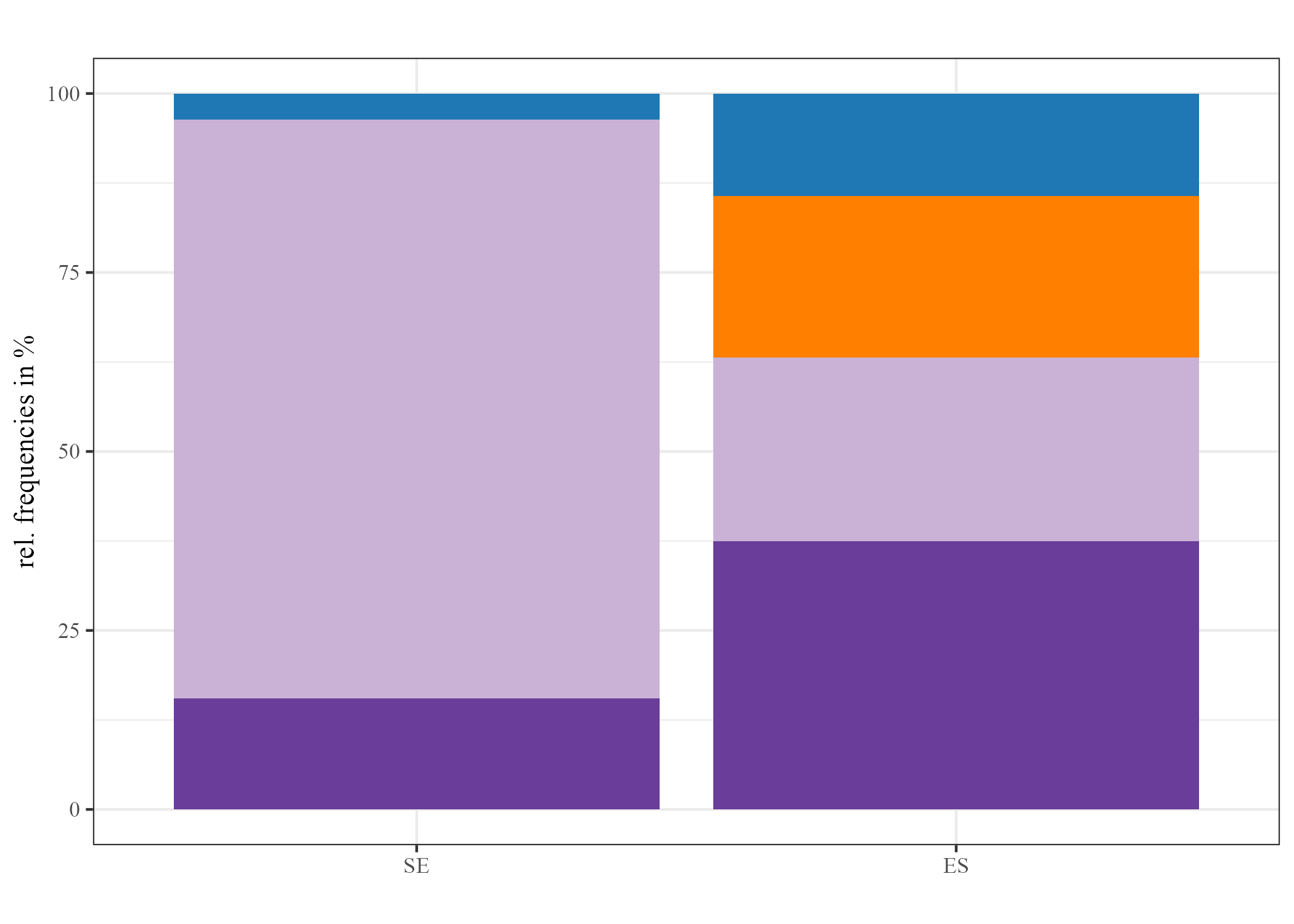}
\caption{German BLOOM 0.350 B}\label{coherence_bloom_0350}
\end{subfigure}
\begin{subfigure}{0.32\textwidth}
\fcolorbox{white}{SkyBlue}{\includegraphics[width = \textwidth]{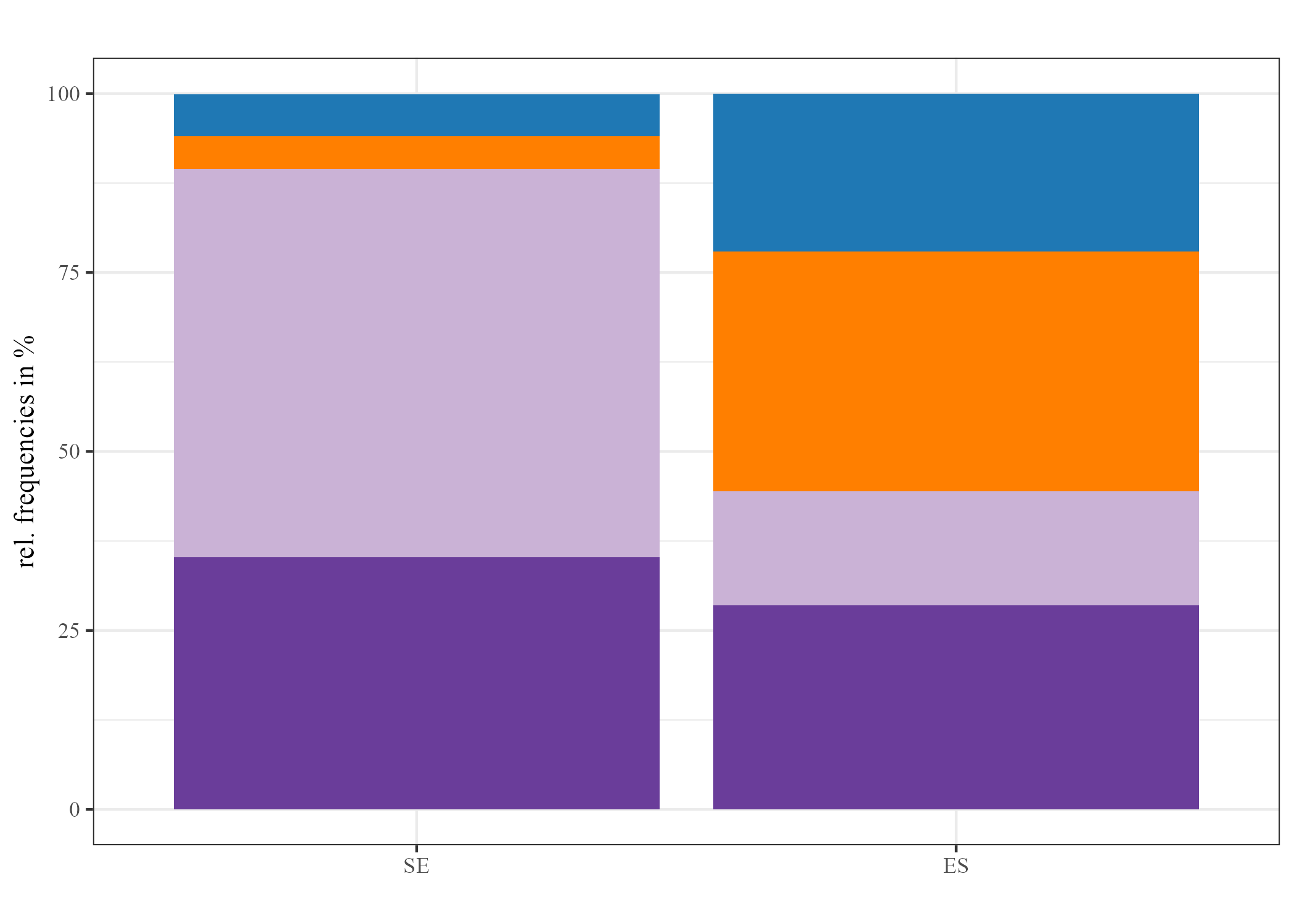}}
\caption{German BLOOM 1.5 B}\label{coherence_bloom_15}
\end{subfigure}
\hspace{1pt}
\begin{subfigure}{0.32\textwidth}
\includegraphics[width = \textwidth]{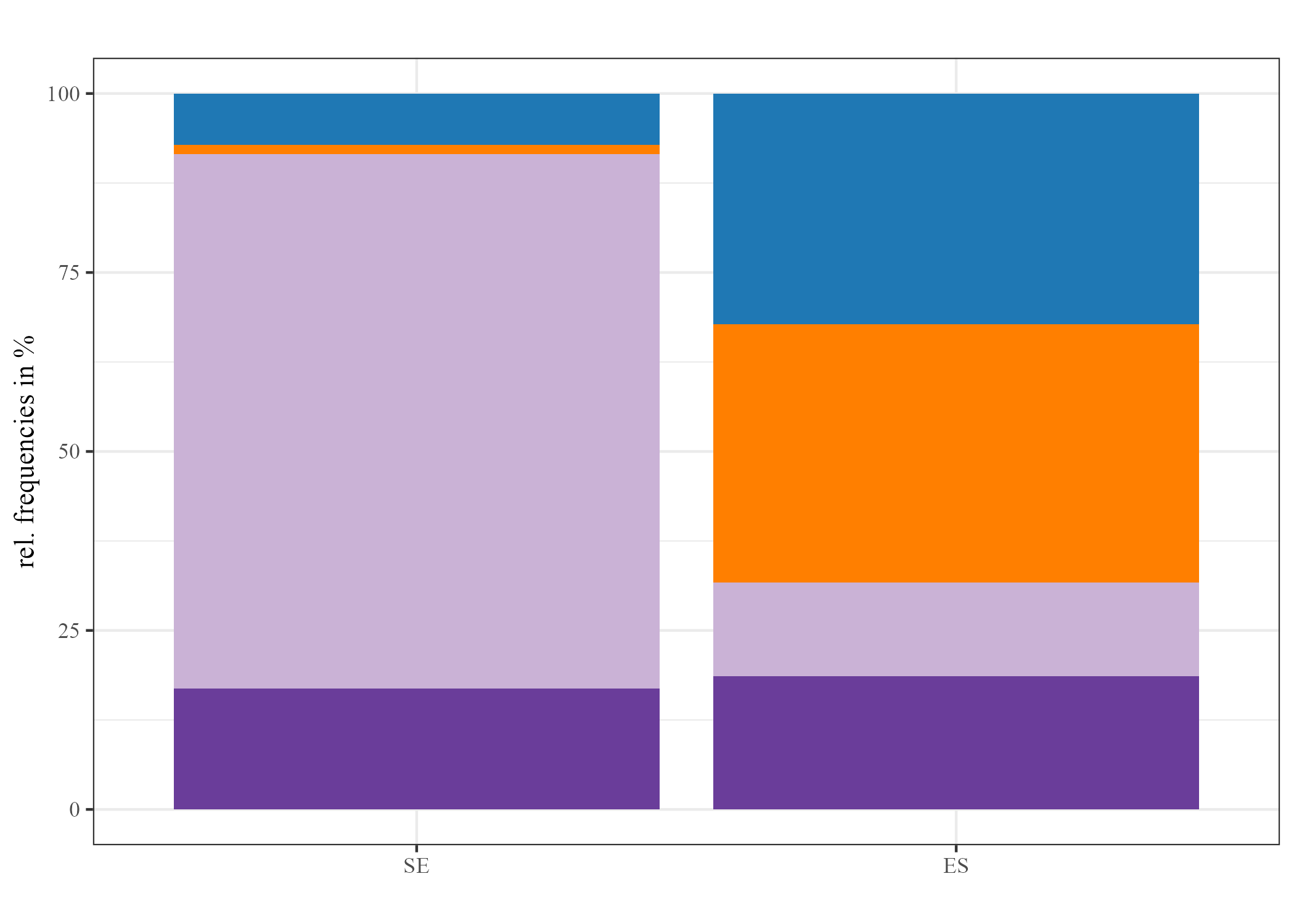}
\caption{German BLOOM 6.4 B}\label{coherence_bloom_64}
\end{subfigure}
\caption{Coherence biases of stimulus-experiencer and experiencer-stimulus verbs (left and right-hand bars, respectively) for human data (Figure \ref{coherence_humans}) and the LLMs. The clearly predominant relation in the human data, Explanations, is always plotted as the bottom-most category in the bars.}\label{coherence_plots}
\end{figure}

In general, the LLMs produced a smaller range of connectives (between 8 and 10 different connectives) than humans (15 different clause-initial connectives). Furthermore, whereas seven different relation types (including the category \textit{other}) were identified in the human data, LLMs produced between 4 and 5 different relation types (Consequence and Elaboration relations, as typically coded by \textit{sodass} `and so' and \textit{indem} `by + gerund', respectively, were not observed in LLMs). The proportions of discourse relations for the human ``gold standard'' data (in the light gray frame) as well as all LLMs are plotted individually for stimulus-experiencer (left-hand bar) and experiencer-stimulus verbs (right-hand bar) in Figure \ref{coherence_plots} on page \pageref{coherence_plots}.%
\footnote{The data were subjected to the same, automatized annotation procedure, which is why the proportion of explanations is lower for the human data than discussed in the introduction of Experiment 2.} %
As evidenced by Figures \ref{coherence_gpt2} through \ref{coherence_bloom_64}, no LLM mirrors the picture characteristic for human data, where significantly more than half of the discourse relations were of explanation type. To the contrary, in all LLMs this proportion was significantly lower than 50\% for both verb classes and in no LLM did the proportion of explanations exceed 37.5\%. Furthermore, four LLMs (among them the multilingual mGPT and XGLM 1.7 B and 4.5 B) produced significantly more explanations for experiencer-stimulus than for stimulus-experiencer verbs, where humans displayed no significant difference (with a numerical difference in the same direction, however). Overall, (monolingual) German Bloom 1.5 B (Figure \ref{coherence_bloom_15}) may be considered to come closest to the human data, with explanations making up 35.2\% of discourse relations for stimulus-experiencer verbs and 28.5\% for experiencer-stimulus verbs.

The clearly dominant relation in LLMs were temporal relations (as marked by the temporally synchronous connective \textit{als} `as'/`when'), in particular following stimulus-experiencer verbs, with values ranging from 93.7\% (Figure \ref{coherence_mgpt}) to 54.3\% (Figure \ref{coherence_bloom_15}). For experiencer-stimulus verbs, the picture is more mixed, with temporal relations only in the majority in one LLM (Figure \ref{coherence_mgpt}) and explanations even turning up as the most frequent relation in two LLMs (Figures \ref{coherence_fb_45} and \ref{coherence_bloom_0350}).

\begin{table}[ht!]
\begin{center}
\setlength\tabcolsep{.2em}
\begin{tabular}{lrrlcrrl}\toprule
    Data & \multicolumn{3}{c}{(Intercept)} & & \multicolumn{3}{c}{\textsc{v.class}} \\
    & \multicolumn{3}{c}{Prop.\ of expl.} & & \multicolumn{3}{c}{Main effect} \\
    \cmidrule(lr){2-4} \cmidrule(lr){5-8}
    & \multicolumn{3}{c}{$z$} & & \multicolumn{3}{c}{$\chi^2(1) =$} \\
    \midrule
    \rowcolor{lightgray}\textsc{human} & \Checkmark & $6.93$ & *** & & \Checkmark & 0.0 &  \\
    G-GPT-2 & \XSolidBrush & $-7.61$ & *** & & \Checkmark & 0.2 &  \\
    mGPT & \XSolidBrush & $-9.45$ & *** & & \XSolidBrush & 10.7 & *** \\
    XGLM 0.564B & \XSolidBrush & $-7.74$ & *** & & \Checkmark & 0.5 &  \\
    XGLM 1.7B & \XSolidBrush & $-8.33$ & *** & & \XSolidBrush & 4.4 & *  \\
    XGLM 2.9B & \XSolidBrush & $-5.88$ & *** & & \Checkmark & 0.0 &  \\
    XGLM 4.5B & \XSolidBrush & $-6.80$ & *** & & \XSolidBrush & 5.0 & * \\
    XGLM 7.5B & \XSolidBrush & $-6.04$ & *** & & \Checkmark & 1.0 &  \\
    G-Bloom 0.350B & \XSolidBrush & $-5.88$ & *** & & \XSolidBrush & 5.0 & * \\
    \rowcolor{SkyBlue} G-Bloom 1.5B & \XSolidBrush & $-3.86$ & *** & & \Checkmark & 2.1 &  \\
    G-Bloom 6.4B & \XSolidBrush & $-6.18$ & *** & & \Checkmark & 0.4 & \\ \bottomrule
\end{tabular}
\end{center}
\caption{Statistics for the coherence bias performance of the different LLMs as compared to human data (from left to right):\  Significance levels indicated by asterisks:\ * = $p < .05$; ** = $p < .01$; *** $= p < .001$. LLM values marked with a \Checkmark{} signify a result in the direction of the human data, while a \XSolidBrush{} indicates a result contrary to the human data (cf. section \ref{sec:stats_exp2}).}\label{inferential:coherence}
\end{table}

A final note on the factor of \textsc{Gender order} is in order. Whereas an effect of this factor was observed for most LLMs in Experiment 1, no systematic pattern was found in Experiment 2.

\section{Experiment 3:\ Anaphoric Form Effects of I-Caus}
\label{sec:exp3}
We now turn to the third bias discussed above, according to which greater proportions of simple anaphoric forms such as personal pronouns are used to establish bias-congruent next-mention coreference than in  bias-incongruent continuations \citep[among others]{FukumuravanGompel2010,KehlerRohde2013,RohdeKehler2014,RosaArnold2017,WeatherfordArnold2021}. This follows a generally assumed principle, according to which more reduced forms are used to refer to more salient \citep{Ariel1990,Gundeletal1993} or even expected referents \citep{Arnold2010}.

The human data for comparison were elicited by \citet{BottSolstad2023} in a forced reference production task \citep{FukumuravanGompel2010} in which participants were asked to provide a continuation about a particular referent to even out the strong coreference biases. The experiments of relevance in the current study applied a $2\times 2$ design with the factors \textsc{grammatical function focus} (\textit{subject} vs.\ \textit{object}) and \textsc{verb class} (\textit{stimulus-experiencer} vs.\ \textit{experiencer-stimulus}):%
\footnote{One of the major results of the study by \citet{BottSolstad2023} was that the effects are more pronounced and consistent for prompts in which same-gender referents are used (e.g., \textit{Mary fascinated Jane because}) thereby increasing the pressure to disambiguate using different referential expresions. In order to be able to work with automatic and error-less annotation in the LLM experiments below, we will restrict our attention to different-gender prompts in the present paper.}

\ex.\label{se_focus} \textbf{Stimulus-experiencer verb \textit{faszinieren} `fascinate'}
\a. \fbox{\textit{Maria}} \textit{fasizinierte Peter, weil \dots} \HandPencilLeft \hfill(subject focus)\\
`Maria fascinated Peter because \dots'\label{ex:subject_focus}
\b. \textit{Maria faszinierte} \fbox{\textit{Peter}}\textit{, weil \dots} \HandPencilLeft \hfill(object focus)\\
`Maria fascinated Peter because \dots'

\ex.\label{es_focus} \textbf{Experiencer-stimulus verb \textit{bewundern} `admire'}
\a. \fbox{\textit{Emma}} \textit{bewunderte Karl, weil \dots} \HandPencilLeft \hfill(subject focus)\\
`Emma admired Karl because \dots'
\b. \textit{Emma bewunderte} \fbox{\textit{Karl}}\textit{, weil \dots} \HandPencilLeft \hfill(object focus)\\
`Emma admired Karl because \dots'

The major categories observed during the annotation of the data were, in increasing order of morpho-syntactic complexity \citep{Ariel2001}, personal pronouns (\textit{er}/\textit{she} `he/she'), demonstrative pronouns (both of \textit{diese/r} as well as \textit{der/die} type) and repeated names \citep[for details, cf.][]{BottSolstad2023}. 

\begin{figure}[h!]
    \centering
    \includegraphics{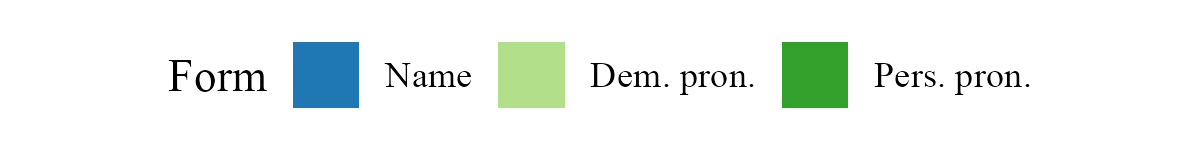}
    \includegraphics[width=0.6\textwidth]{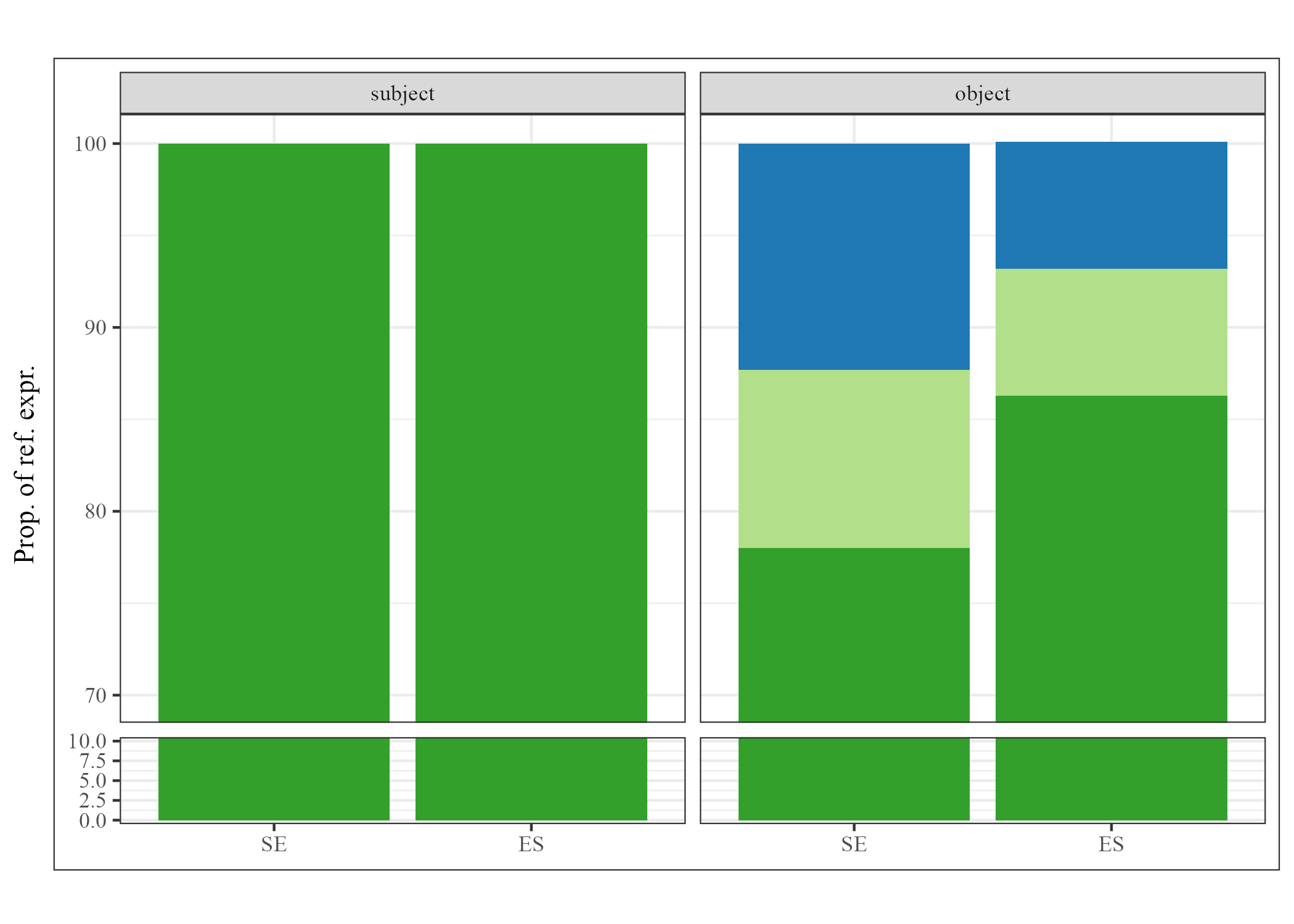}
    \caption{Anaphoric forms produced in \citet{BottSolstad2023}. Data pooled together from their Pilot Study (subject reference conditions with antecedents of different gender, left-hand side) and their Exp.\ 2 (object reference conditions with different gender referents, right-hand side). Note: ES = experiencer-stimulus, SE = stimulus-experiencer, Name = Proper name, Dem.\ pron.\ =  demonstrative pronoun, Pers.\ pron.\ = personal pronoun}
    \label{Figure_forms_human}
\end{figure}

Figure \ref{Figure_forms_human} presents combined data from two experiments reported in \citet[][Pilot Study \& Exp. 2]{BottSolstad2023}. Two lines of result are of relevance in the context of our test case for LLMs. On the one hand, the results display the well-known effect of \textsc{grammatical function}:\ The proportion of simpler forms is larger for coreference to the subject than to the object \citep{Crawleyetal1990}. In addition, the experiments an interaction between \textsc{grammatical function} and \textsc{verb class}:\ While coreference with the subject of the prompt sentence was uniformly established with personal pronouns regardless of bias congruency, next-mention bias modulated the forms used for coreference to objects. Participants produced slightly more pronouns than other forms in bias-congruent continuations than in bias-incongruent ones (GLMER analyzing likelihood to refer by a personal pronoun:\ $\chi^2(1) = 6.97$; $p<.01$).

\subsection{Methods}

The design of this experiment necessarily deviated from the one in \citet{BottSolstad2023}, since it was not possible to prompt the models to adhere to the forced referent paradigm applied in that study.\footnote{Remember that the LLMs used were exclusively foundation models that are not trained to follow instructions. As such, prompting the models to complete the IC prompts in a certain way -- especially a more complex linguistic restriction like constructing an anaphoric reference -- has very little effect on the actual model output.} Instead, the forced-reference procedure was emulated in the implementation of the experiment. The technical setup was as follows:

With the antecedents differing in gender, the subject and object focus conditions were realized using a constrained beam search decoder and a custom constraint function, effectively constraining the the LLMs output to produce a continuation in which the first word would be gender-congruent with the subject or object, respectively. Only three forms were allowed: Personal pronouns (\textit{er} `he' or \textit{she} `sie'), proper names and demonstratives (\textit{dieser} `this' [masc.] and \textit{diese} `this' [fem.]). For instance, subject focus in \ref{ex:subject_focus} was achieved by constraining the possible tokens the language model was allowed to generate immediately following the prompt to only \textit{sie} `she', \textit{Mary} and \textit{diese} `this one (fem.)'.

We generated this data until a sufficiently large number of bias-congruent as well as bias-incongruent continuations were generated. This allowed us to run comparable statistical analyses as for the human data. Sampling continued until a total of 8,000 continuations were generated, 1,000 in each condition.

\subsection{Data annotation and selection}

The procedure for annotating the data ensured that every continuation submitted to statistical analysis could be assigned a syntactic parse. No other criteria for acceptability were applied. Only very few data points had to be excluded (0.1\% and 0.2\% in two LLMs, respectively). We once again checked our automatic annotater against the human-annotated data from \citet{BottSolstad2023} and found an agreement of $\kappa = .74$ for anaphoric forms. Restricting the agreement calculations to only the sentences deemed sensible by their human annotator raised the score to  $\kappa = .94$.

\subsection{Statistical analysis}

Like in the previous experiments, we fitted mixed-effects binomial logistic regression models \citep{Batesetal2015} for each LLM. This time, the binomial dependent variable coded whether the first referring expression was a personal pronoun or not, analogous to the human data. The model included the fixed effects of \textsc{verb class} and (the counter-balancing factor) \textsc{gender order} in addition to their interaction. The random effects structure included by-verb random slopes for \textsc{gender} order in addition to a random intercept for verbs. Beyond this, we proceded as described in section \ref{sec:analysis:coref} with respect to model comparison.

\subsection{Results and Discussion}

\begin{figure}[h!]
\begin{center}
\includegraphics{graphs/forms_icaus/icaus_form_legend.png}
\end{center}
\begin{subfigure}{0.32\textwidth}
\fcolorbox{white}{black}{
\includegraphics[width = .96\textwidth]{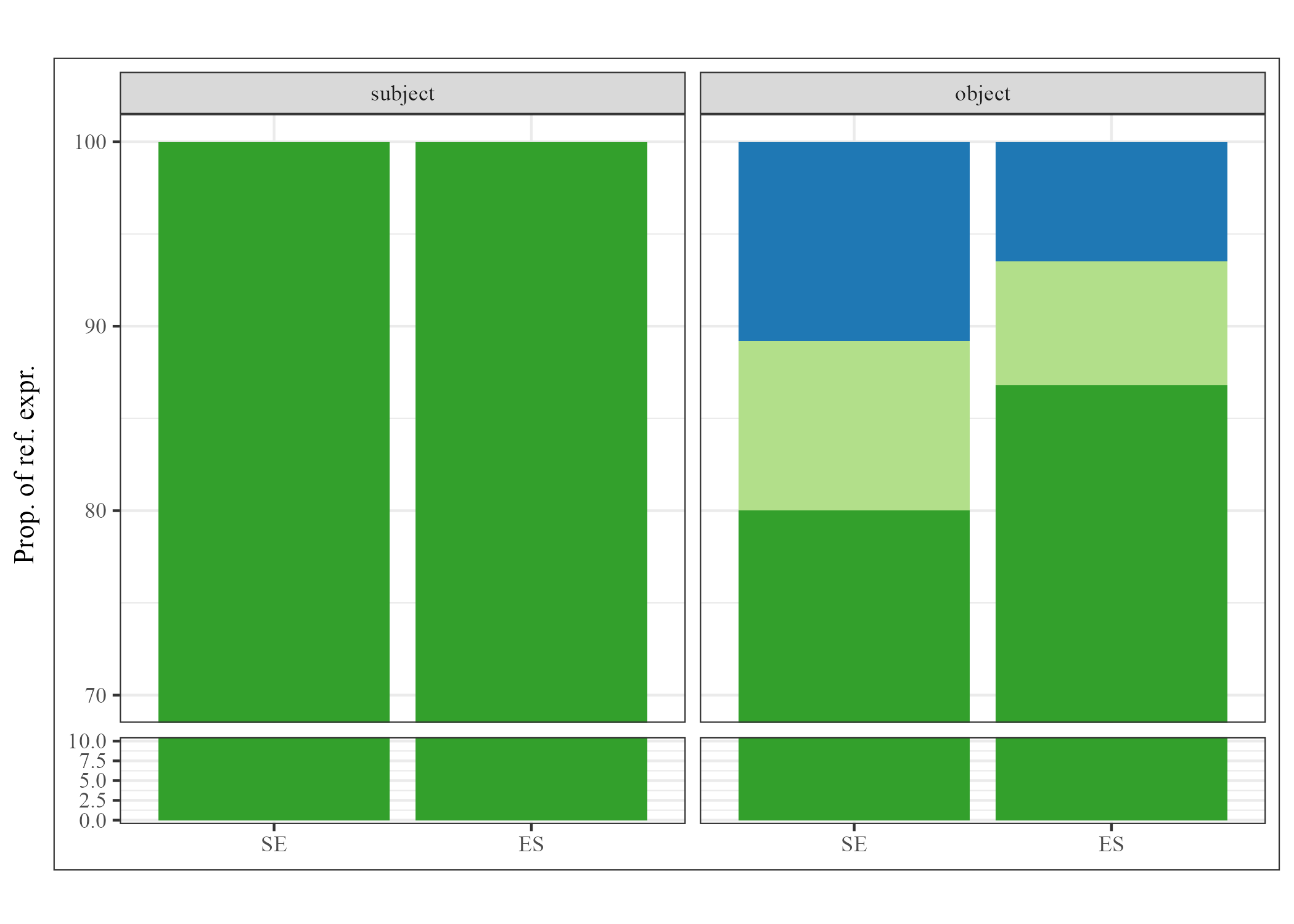}}
\caption{Human data (machine coded)}\label{forms_icaus_humans}
\end{subfigure}
\hspace{0pt}
\hspace{0pt}
\begin{subfigure}{0.32\textwidth}
\fcolorbox{white}{SkyBlue}{\includegraphics[width = \textwidth]{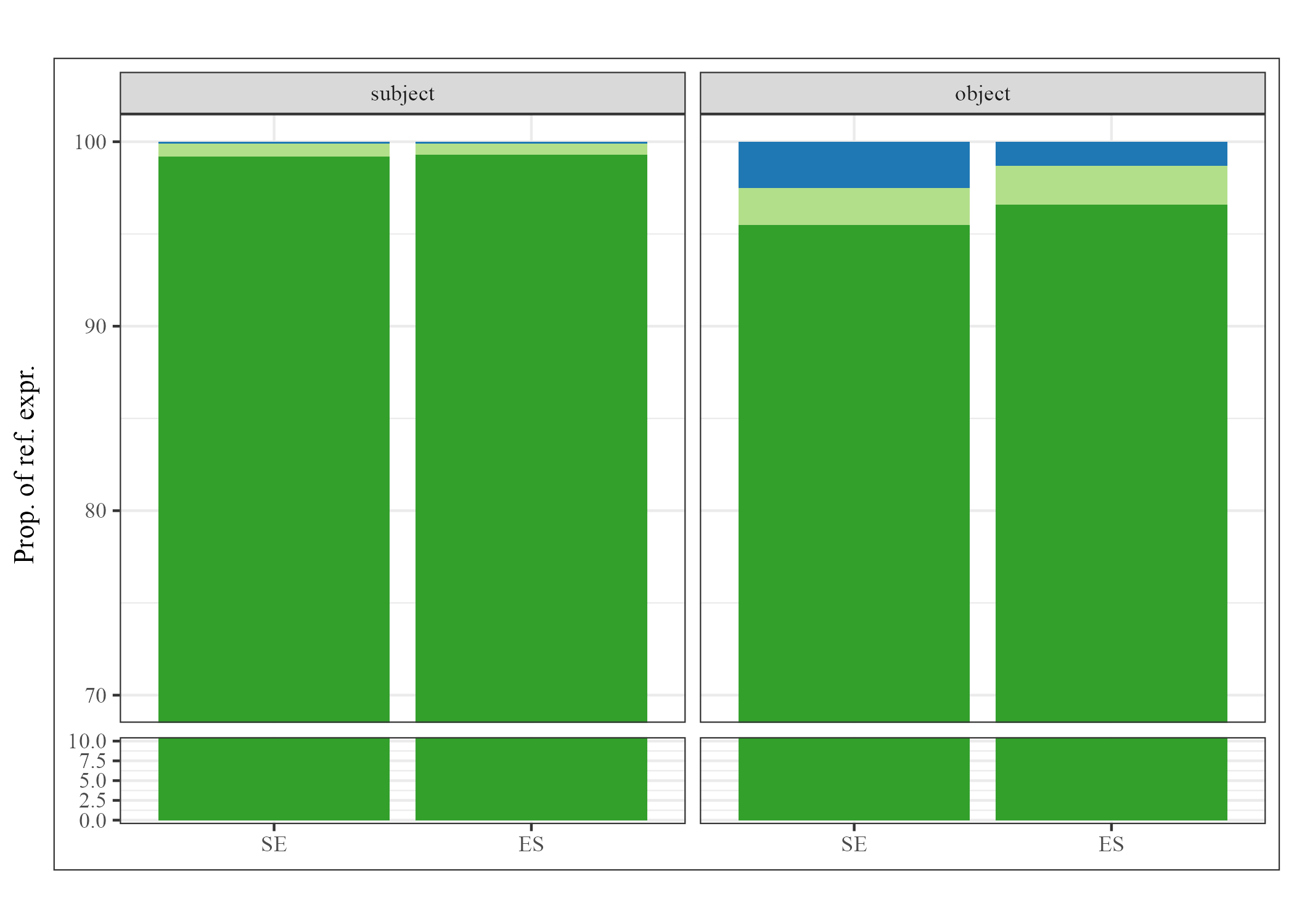}}
\caption{German GPT-2}\label{forms_icaus_gpt2}
\end{subfigure}
\begin{subfigure}{0.32\textwidth}
\includegraphics[width = \textwidth]{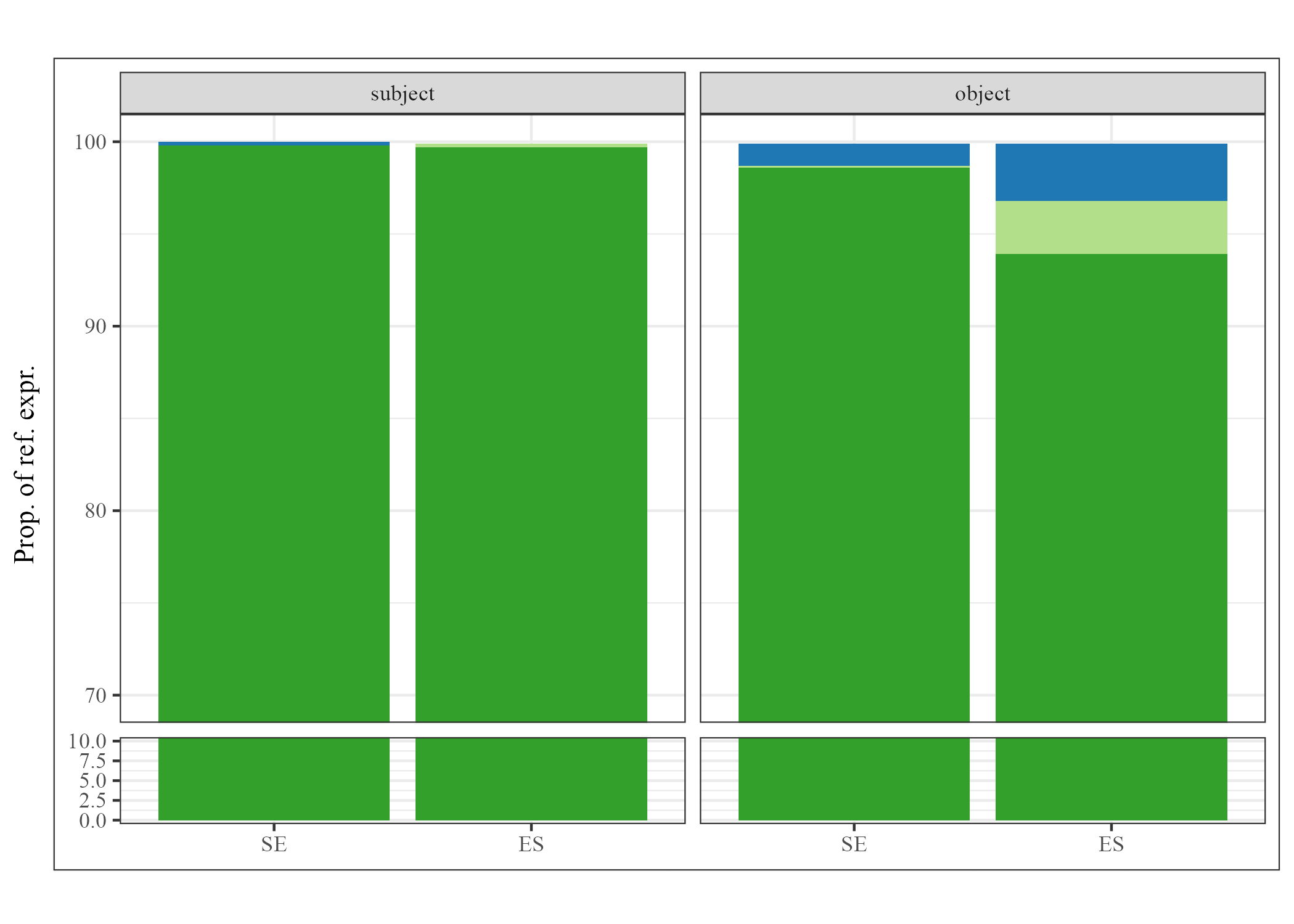}
\caption{XGLM 7.5 B}\label{forms_icaus_fb_75}
\end{subfigure}
\caption{Distribution of anaphoric expressions for I-Caus}\label{forms_icaus_plots}
\end{figure}

As the results were uniformly deficient with regard to coreference bias, we have chosen only to include the data from two LLMs next to the human data in Figure \ref{forms_icaus_plots}. That being said, all LLMs displayed a significant general grammatical function focus effect to the extent that a greater proportion of personal pronouns, that is, the maximally simple referring form in German, was used to establish coreference with the subject than with the object (regardless of verb class bias), as can be seen in Figures \ref{forms_icaus_gpt2} and \ref{forms_icaus_fb_75}. This is an effect consistently found in human data, as discussed above. With regard to effects of I-Caus coreference bias for coreference to the object, however, all LLMs failed to display an effect comparable to that found in human data. In fact, some LLMs even showed an effect in the opposite direction, that is, more personal pronouns for coreference to bias-incongruent arguments, that is, for reference to the object argument of stimulus-experiencer verbs, as illustrated by XGLM 7.5B in Figure \ref{forms_icaus_fb_75}. Only two monolingual models, among them German GPT-2 in Figure \ref{forms_icaus_gpt2} displayed an effect that trended \textit{numerically} towards the effect found in human data, see Figure \ref{forms_icaus_humans}, but not significantly so ($\beta = -0.18, \mathit{SE} = 0.27, z = -0.66$).

Two further remarks are in place on findings that go counter to what is typically found in studies involving human participants. On the one hand, five of the LLMs (among them German GPT-2 in Figure \ref{forms_icaus_gpt2}) produced demonstrative pronouns for expressions coreference to the subject (in the range of 0.4 to 2.7\%), something which is only found in information-structurally highly marked continuations for human beings \citep{BoschHinterwimmer2016}. On the other hand, some LLMs generate a greater proportion of personal pronouns for coreference to the subject of stimulus-experiencer than to that of experiencer-stimulus verbs. Recall that in the human data, only personal pronouns were produced for referring expressions coreferent with the subject argument, regardless of verb class. Finally, it may be noted that, again, no consistent effect of \textsc{gender order} was found for the LLMs (half of the LLMs displayed such an effect, cutting across size and mono-/multilingual properties).

\section{Experiment 4:\ Anaphoric Form Effects of I-Cons}
\label{sec:exp4}

It may not be particularly surprising that no effects of I-Caus coreference bias on anaphoric expressions could be observed in LLMs. After all, only one LLM displayed a significant difference in I-Caus bias whatsoever in Experiment 1. And even this bias showed more variance than that found in human data. For this reason, we also conducted an experiment investigating a possible form effect for I-Cons coreference bias, as this effect was found to differ significantly between verb classes in Experiment 1, with selected LLMs displaying almost human-like bias levels. We have no human data for comparison, but again we expect to find a target effect for coreference to the subject vs.\ object and also, if at all, only a coreference bias effect for coreference to the object (notwithstanding the subject congruency effects briefly discussed in Experiment 3).

\subsection{Methods}

The design of this experiment followed the one in Experiment 3, see examples \ref{se_focus} and \ref{es_focus}, with the exception that \textit{weil} `because' was replaced by \textit{sodass} `and so'  to evoke I-Cons biases. Sampling was conducted in the same way as in Experiment 3, resulting in  a total of 8,012 continuations.

\subsection{Data annotation and selection}

The procedure for annotating the data ensured that every continuation submitted to statistical analysis could be assigned a syntactic parse. No other criteria for acceptability were applied. Only very few data points had to be excluded (0.1\% in one LLM).

\subsection{Statistical analysis}

The statistical analysis was parallel to the one in Experiment 3.

\subsection{Results and Discussion}

\begin{figure}[h!]
\begin{center}
\includegraphics{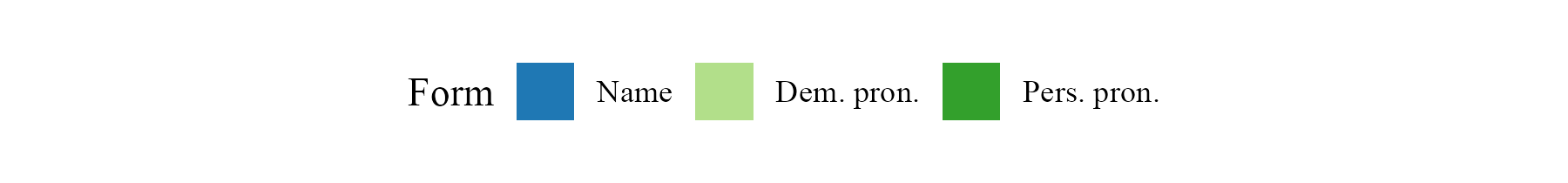}
\end{center}
\begin{subfigure}{0.32\textwidth}
\includegraphics[width = \textwidth]{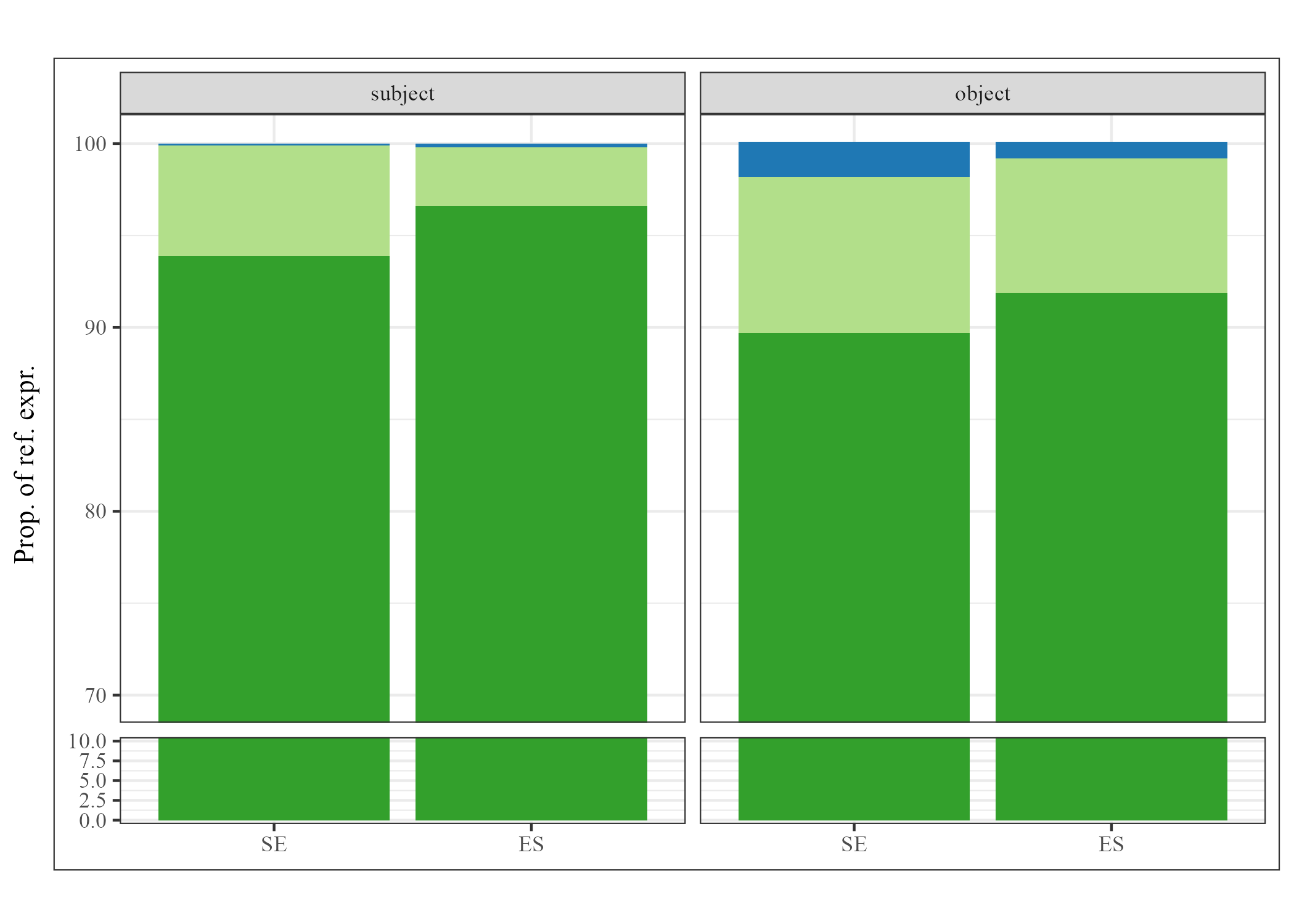}
\caption{German GPT-2}\label{fdorms_icons_gpt2}
\end{subfigure}
\begin{subfigure}{0.32\textwidth}
\includegraphics[width = \textwidth]{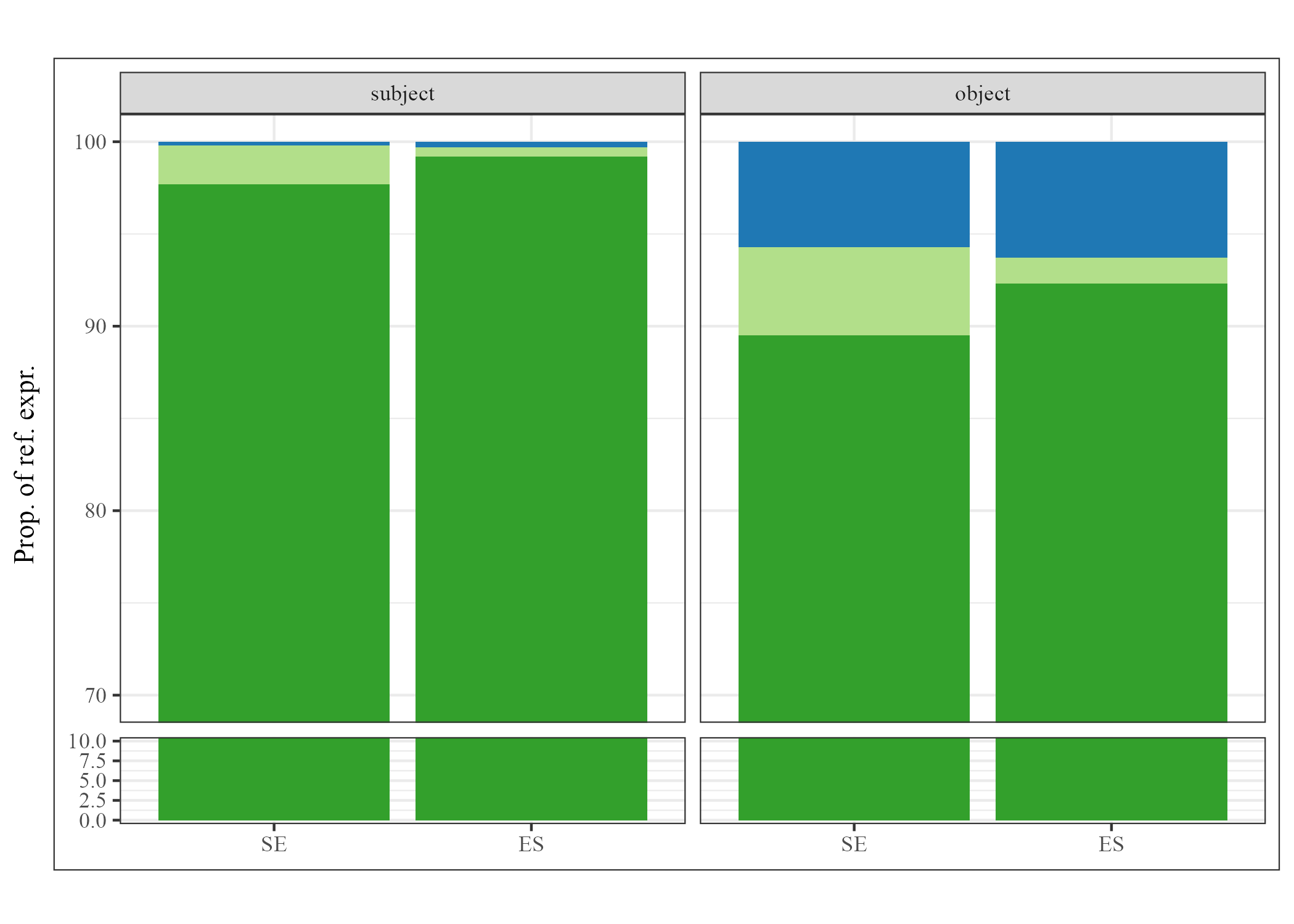}
\caption{XGLM 7.5 B}\label{forms_icons_fb_75}
\end{subfigure}
\begin{subfigure}{0.32\textwidth}
\includegraphics[width = \textwidth]{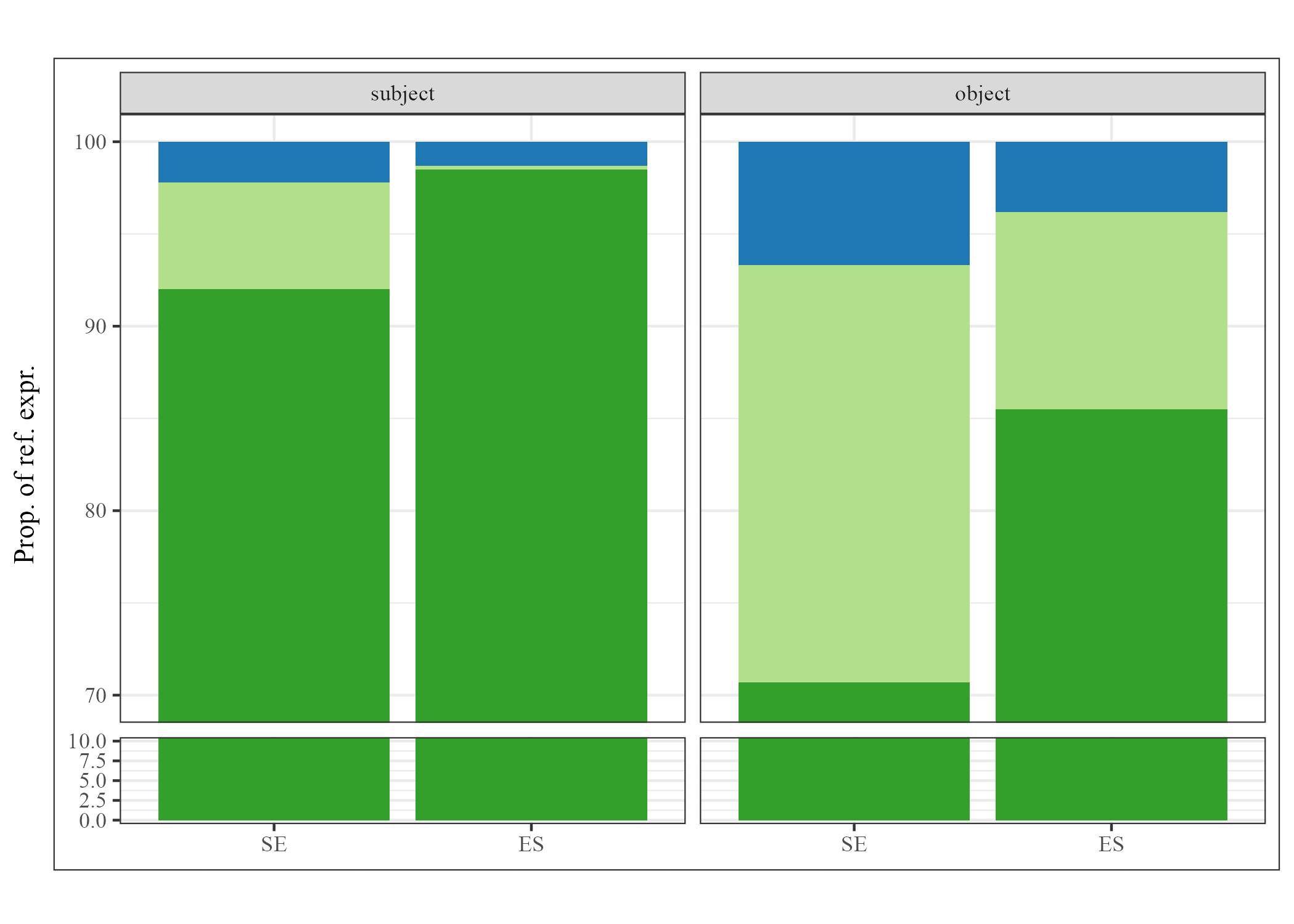}
\caption{German BLOOM 1.5 B}\label{forms_icons_bloom_15}
\end{subfigure}
\hspace{1pt}

\begin{subfigure}{0.32\textwidth}
\fcolorbox{white}{SkyBlue}{\includegraphics[width = \textwidth]{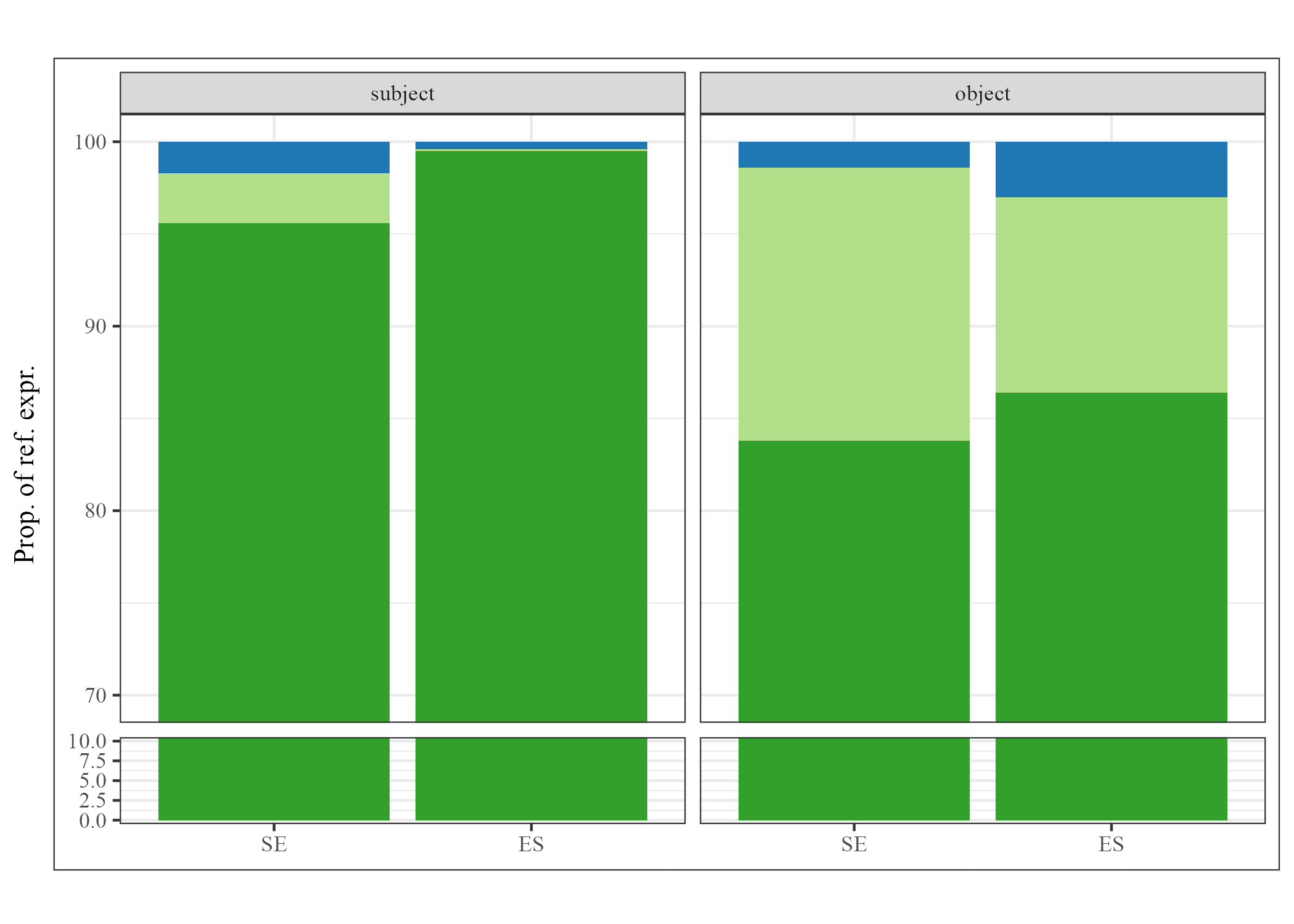}}
\caption{German BLOOM 6.4 B}\label{forms_icons_bloom_64}
\end{subfigure}

\caption{Distribution of anaphoric expressions for I-Cons}\label{forms_icons_plots}
\end{figure}

The results for selected LLMs are presented in Figure \ref{forms_icons_plots}. The grammatical function focus effect from Experiment 3 was replicated for I-Cons:\ A greater proportion of personal pronouns were found for referential expressions coreferent with the subject than the object in all LLMs (Figures \ref{forms_icons_fb_75} and \ref{forms_icons_bloom_15}). For coreference to the object, allmost all LLMs showed no significant effect. However, as for I-Cons, almost all models displayed a trend in a direction which would be unexpected based on human data:\ A greater proportion of personal pronouns was found for incongruent continuations, that is for experiencer-stimulus verbs. In two LLMs, German Bloom 350M and German Bloom 1.5B (Figure \ref{forms_icons_bloom_15}), this effect even reached significance. On the other hand, only one LLM showed at least a balanced destribution for object reference across verb classes (German Bloom 6.4B in Figure \ref{forms_icons_bloom_64}). It may be noted that only two LLMs displayed an effect for \textsc{gender order} (XGLM 4.5 B and German GPT-2).

Again, some further interesting deviations from human data could be observed. First, all LLMs displayed an effect of \textsc{verb class} for refererring expressions to the subject:\ A greater proportion of personal pronouns was found for congruent (experiencer-stimulus) than incongruent (stimulus-experiencer) continuations. Based on human data, we would expect no effect of \textsc{verb class} whatsoever for subject targets. Second, all LLMs produced a considerable amount of demonstrative pronouns for anaphoric expressions coreferent with the subject (range:\ 1.4 to 7.2\%). To reiterate, human participants typically only do so in highly particular information-structural configurations \citep{BoschHinterwimmer2016}.

\section{General Discussion}
\label{sec:discussion}

In our four experiments, we compared data generated with mono- and multilingual LLMs spanning a range of model sizes with data provided by human participants in an experimental setting. We did so for the realm of Implicit Causality verbs, for which psycholinguistic research has found participants to display biases with regard to three phenomena:\ (i) the establishment of coherence relations, (ii) of coreference relations, and (iii) the use of particular referring expressions. In doing so, we treated this triad of biases as a benchmark to assess the capabilities of LLMs in demonstrating human-congruent biases, which serve as a robust proxy for more general discourse understanding capabilities.

In Experiment 1, investigating coreference biases, after prompts like \textit{Mary fascinated/admired Peter because/and so \dots{}}, we found only the largest monolingual LLM (German Bloom 6.4B) to display an I-Caus bias (after \textit{weil} `because') that significantly trends towards the bias found in humans. However, almost all LLMs displayed the converse I-Cons bias (after \textit{sodass} `and so'). We will return to this asymmetry below.

In Experiment 2, we investigated the coherence bias, comparing the connectives humans and LLMs were most likely to use after prompts ending in a comma:\ \textit{Mary fascinated/admired Peter, \dots{}}. While we found explanation biases of around 80\% for the two verb classes in humans, no LLM displayed an explanation bias above 50\%. Instead, LLMs typically produced temporal relations. Again, a monolingual LLM came closest, with explanations constituting the major category for at least one of the verb classes (German Bloom 1.5B). Interestingly, although of the same family as the model with the best performance in the coreference bias experiment, this model was significantly smaller than the `victoriuos' LLM for coreference.

Finally, in Experiments 3 and 4, we investigated whether LLMs show a bias towards producing simpler referring expressions (the pronoun \textit{she} as opposed to the proper name \textit{Mary}) for biased arguments after \textit{weil} `because' and \textit{sodass} `and so' prompts. No LLM displayed any such effect for I-Caus in Experiment 3, which may not be all that surprising since the majority of models displayed no I-Caus coreference bias whatsoever. However, even for I-Cons bias, which the LLMs mostly did display, no bias effect on referring expression was found. It should be noted, however, that we were able to replicate psycholinguistic findings according to which participants use a higher proportion of simpler forms, that is, personal pronouns, to refer to subject arguments as opposed to object arguments, regardless of bias status.

One notable observation from Experiment 1 is the difference between I-Caus and I-Cons biases across all models. In general, the I-Cons bias was not only more pronounced (in the human-like direction), but also less noisy than the causality bias. This indicates a stronger understanding of the concept of consequentiality by the models, as they were more reliably able to capture and express an associated discourse bias. Under this assumption, causality requires a stronger functional linguistic competence \citep[as described by][]{mahowald2024dissociating}, whereas consequential discourse relations -- like grammar -- are moreso governed by formal linguistic competence. This assumption aligns with the two-mechanism theory for IC of \citet{BottSolstad2021}, wherein I-Caus biases are governed by a special semantic mechanism associated with IC verbs. Consequentiality biases on the other hand arise from a broader contiguity principle \citep{Kehler2002} that is not unique to Implicit Causality, but applies to discourse in general. While the latter can be more easily generalized by an LLM, the former requires a robust functional understanding of language and causality.

This assumption would also explain the relatively low proportion of explanation relations formed by all models in Experiment 2, as \citet{BottSolstad2021} attribute both the coreference as well as coherence causality bias to the same mechanism. However, this behaviour might also be a statistical frequency effect of the training data. \citet{AsrDemberg2012} found that causal discourse relations in the PDTB corpus were mostly formed implicitly without a discourse marker, whereas temporal relations are mostly formed through the use of an explicit connective. If this pattern extends to the training data of the tested LLMs, it could explain the abundance of temporal relations, as our experimental design forced the generation of a connective to form sensible continuations.

In general, no definite conclusions about model behaviour can be drawn from our data, as all LLMs performed poorly and inconsistently across the board. Understanding the stark divergence between human and model biases would require seperate analyses. Still, our findings suggest that midrange models (up to ~7 billion parameters) fail to capture humanlike IC coherence biases, shying away from causality in favour of more linear temporal constructions (Experiment 2). Similarly, they do not have a robust grasp of coreference biases in causal IC contexts (but can capture consequential biases, albeit less pronounced than humans, Experiment 1). They are, however, completely insensitive to anaphoric form biases (Experiments 3 and 4). Although slim, there is a difference between the adherence to the human biases in Experiment 1 compared to Experiments 3 and 4, as predicted by the model shown in Figure \ref{figure:model}. This suggests that IC effects exhibited by the LLMs may, in fact, be governed by a hierarchical structure, meaning that stronger coreference and coherence biases are required for anaphoric form biases to develop.

This data highlights the applicability of our experiments as an interpretable (hierarchical) discourse benchmark for language models. It can be used to compare models with stronger (functional) linguistic competence, as some of the discourse biases presented in this paper are completely inaccessible to some LLMs with 7 billion parameters. In particular, there is not only a quantitative difference between our observations and the IC biases exhibited by ChatGPT in \cite{cai2023does}, but also a qualitative one. In their study, ChatGPT corresponds to human I-Caus coreference biases and does so strongly, even eclipsing human bias, resulting in behaviour that is maybe just as inhuman as our observations, only in the opposite direction. Since the language model behind ChatGPT is inaccessible, the exact causes for this difference cannot be determined. We did not find any effect of model size in our data, however, with even larger parameter counts such as in ChatGPT, a measurable effect might appear. Alternatively, the emergence of a strong bias in ChatGPT could be caused by instruction tuning, a feature not present in any of the LLMs we tested. A third possibility, although unlikely, is a difference in languages, as \citet{cai2023does} conducted their experiments in English.

These questions can be addressed by running our experiments on other language models of larger sizes, including instruction-tuned models, and expanding our benchmark to other languages.%

\section{Conclusions}
\label{sec:conclusion}

In this paper, we constructed a benchmark for assessing Implicit Causality biases on three levels of discourse, based on psycholinguistic experiments with gold-standard human comparison data. We used this benchmark to investigate two main questions: (i) whether LLMs exhibit similar discourse biases to humans in the context of Implicit Causality, and (ii) whether these biases differ in larger vs.\ smaller and mono- vs.\ multilingual LLMs. Our experiments showed that none of the models analyzed (German GPT-2, a German BLOOM model family, mGPT, the XGLM model family) expressed humanlike IC biases, ranging from only a slight bias, to no bias or even a bias in the wrong direction. Although some models performed well in some tasks, this behaviour was not consistent across experiments, with no apparent pattern to be found. However, none of the models were even slightly sensible to IC bias in the context of anaphoric forms, suggesting that part of our benchmark may be too subtle for the LLMs investigated in this paper.  

In order to fully address the second research question, our benchmark must thus be expanded to models with stronger linguistic capabilities. In particular, future work should seek to bridge the gap between our observed data and the human-like behaviour of ChatGPT proported by \citet{cai2023does}. To this end, the effects of even larger model size as well as instruction-tuning need to be analyzed.

In a similar vein, future work should investigate the causes for the discrepancy in model behaviour. Our observations that models struggled more with causality than consequentiality and select temporal discourse relations as default -- as opposed to causal relations in humans -- merit an analysis whether this is a function of corpus training data or model architecture.

\label{lastpage}

\end{document}